\pdfoutput=1
\documentclass[11pt]{article}
\DeclareUnicodeCharacter{202F}{\,}

\usepackage[final]{acl}
\usepackage{algorithm}
\usepackage{algpseudocode}
\usepackage{subcaption}
\usepackage{adjustbox}
\usepackage{makecell}  
\usepackage[table]{xcolor}
\usepackage{colortbl}
\usepackage{tabularx}
\usepackage{booktabs}
\usepackage{array}
\definecolor{rocheblue}{HTML}{005EB8}
\usepackage{times}
\usepackage{latexsym}
\usepackage{graphicx}
\definecolor{rocheblue}{HTML}{005EB8}
\definecolor{rochelightblue}{HTML}{CCE5FF}
\definecolor{rochegray}{HTML}{F0F0F0}
\usepackage{amsmath} 
\usepackage{booktabs}
\usepackage{hyperref}
\usepackage{url}
\usepackage{amssymb} 
\usepackage{xcolor}
\usepackage{hyperref}       % hyperlinks
\usepackage{url}            % simple URL typesetting
\usepackage{booktabs}       % professional-quality tables
\usepackage{amsfonts}       % blackboard math symbols
\usepackage{nicefrac}       % compact symbols for 1/2, etc.
\usepackage{array}
\usepackage{booktabs}
\usepackage{xcolor}
\usepackage{colortbl}
\usepackage{graphicx}
\usepackage{multirow}
\usepackage[table]{xcolor}
\usepackage{booktabs}
\usepackage{longtable}
\usepackage{todonotes}
\usepackage{booktabs}
\usepackage{subcaption}
\usepackage[table]{xcolor}
\usepackage{multirow}
\usepackage{colortbl}
\usepackage{graphicx}
\usepackage{booktabs}
\usepackage{geometry}

\definecolor{tableheadbg}{HTML}{F0F6FF}
\definecolor{titleblue}{HTML}{0B57D0}

\definecolor{Industry Proprietary  Datablue}{HTML}{0066CC}
\definecolor{lightblue}{HTML}{EAF4FF}
\definecolor{tablehead}{HTML}{D9E2F3}

\usepackage[T1]{fontenc}
\usepackage[utf8]{inputenc}

\usepackage{microtype}

\usepackage{inconsolata}

\usepackage{graphicx}
\usepackage[normalem]{ulem}  % Add this in the preamble

\title{From Understanding to Engagement: Personalized pharmacy Video Clips via Vision Language Models (VLMs)}

\author{
  \textbf{Suyash Mishra\textsuperscript{a}},  
  \textbf{Qiang Li\textsuperscript{b}},
  \textbf{Anubhav Girdhar\textsuperscript{c}},
  \textbf{Srikanth Patil\textsuperscript{c}}
\\ 
  \textsuperscript{a}Roche,
  \textsuperscript{b}Accenture,  
  \textsuperscript{c}Involead,
\\
  \small{
    \textbf{}  
    \href{mailto:suyash.mishra@roche.com}{suyash.mishra@roche.com},
    \href{mailto:qiang.i.li@accenture.com}{qiang.i.li@accenture.com},
    \href{mailto:anubhav.girdhar@involead.com}{anubhav.girdhar@involead.com},
    \href{mailto:srikanth.patil@involead.com}{srikanth.patil@involead.com} 
    }
 }

\begin{document}
\maketitle

\begin{abstract}
%\todo{What is the business use that that it is used / can be used for?}
%Vision Language Models (VLMs) offer transformative potential for digitizing and enhancing pharmacyceutical industry. Traditional manual tagging and captioning of multi-modality data as text, images, video, audio, and web links can lead to errors, poor quality, and inefficiencies in content reuse. Given the enormous volume of video and audio data, manual labeling is highly labor-intensive and incurs significant costs. Leveraging GenAI, particularly VLM and ALM in a real industry large-scale production system, which analyzes content and automatically generates captions along with relevant tags, in turn supports natural language search and delivers highly relevant results. 

Vision Language Models (VLMs) are poised to revolutionize the digital transformation of pharmacyceutical industry by enabling intelligent, scalable, and automated multi-modality content processing. Traditional manual annotation of heterogeneous data modalities (text, images, video, audio, and web links), is prone to inconsistencies, quality degradation, and inefficiencies in content utilization. The sheer volume of long video and audio data further exacerbates these challenges, (e.g. long clinical trial interviews and educational seminars). 

Here, we introduce a domain-adapted Video-to-Video Clip Generation framework that integrates Audio-Language Models (ALMs) and Vision Language Models (VLMs) to produce highlight clips. Our contributions are threefold: (i) a reproducible Cut \& Merge algorithm with fade-in/out and timestamp normalization, ensuring smooth transitions and audio/visual alignment; (ii) a personalization mechanism based on role definition and prompt injection for tailored outputs (marketing, training, regulatory); (iii) a cost-efficient e2e pipeline strategy balancing ALM/VLM-enhanced processing. 
Evaluations on Video-MME benchmark (900) and our proprietary dataset of 16,159 pharmacy videos across 14 disease areas demonstrate 3–4× speedup, 4× cost reduction, and competitive clip quality. Beyond efficiency gains, we also report our methods improved clip coherence scores (0.348) and informativeness scores (0.721) over state-of-the-art VLM baselines (e.g., Gemini 2.5 Pro), highlighting the potential of transparent, custom extractive, and compliance-supporting video summarization for life sciences. \href{https://video-clips-highlight-generator-338849523617.us-west1.run.app/}{Demo access}.

%Unlike prior work focusing on generic video summarization, we propose a domain-adapted end-to-end pipeline with novel contributions: (i) a reproducible Cut \& Merge algorithm, plus an end-to-end pipeline, ensuring clips smooth transitions and audio/visual alignment, and (ii) a personalization strategy leveraging role-aware prompting to generate tailored clinical/educational/promotional video clips. Our video-to-video architecture enables scalable, cost-efficient, infinite video-to-video processing.  Experiments conducted on the Video-MME benchmark \citep{fu2024video} and our pharmacyceutical proprietary dataset, our methods have achieved a 3-4 times increase in processing speed and 4 times reduction in cost, comparable coherence scores (0.348) as Gemini 2.5 Pro model methods (0.446) and top informativeness scores 0.721 than Gemini 2.5. Our framework offers a practical, scalable business solution for managing large amounts of video data (e.g. >21K) under similar GPU constraints, facilitating the creation of video derivatives. We also provide access to our \href{https://video-clips-highlight-generator-338849523617.us-west1.run.app/}{Demo}.

$\overline{\text{{* Patent application submitted to the EPO.}}}$
%\todo{Add main contributions more detailed here. Abstract can always be shortened afterwards, let's view it as executive summary for now. } -> we add in intro section and details on appendix sections
%(e.g., MP4, M4V, QuickTime, WMV, WebM, MSVideo, MPG, and 3GPP)
\end{abstract}

\begin{figure*}[ht]
    \centering
    \begin{minipage}{1\textwidth}
        \centering
        \includegraphics[width=1\textwidth]{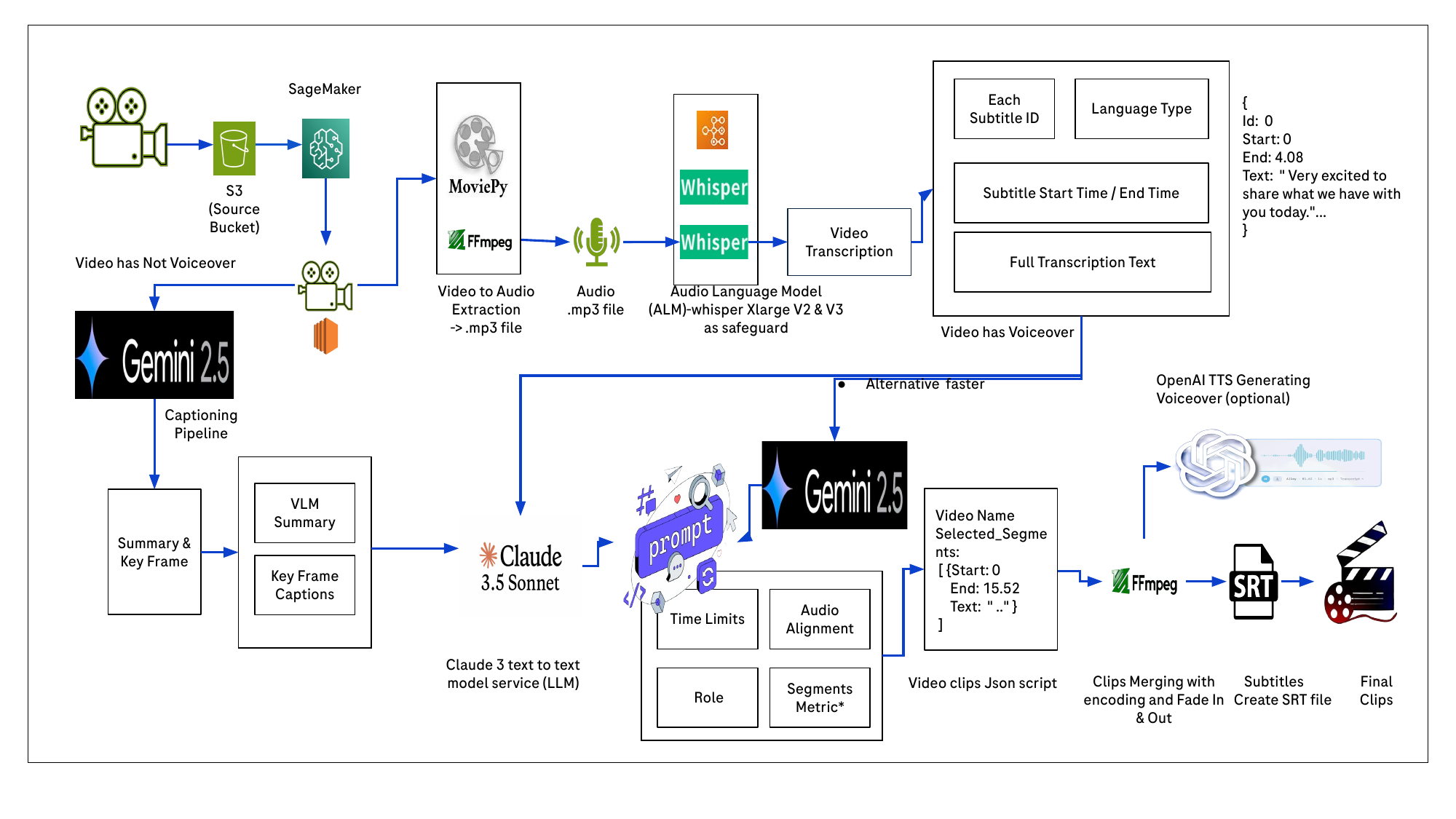} 
    \end{minipage}%
    \vspace{-15pt}
    \caption{Solution architecture blueprint of the underlying LLM/VLM tech stack for video clip generation.} %It successfully enabled content search for 25,326 videos, 888 multilingual audios covering > 20 languages.}
    \label{fig: architecture}
\end{figure*}

\section{Introduction}

In contemporary digital content landscape, efficient management and understanding of video assets are paramount, particularly within specialized domains such as medical and pharmacyceutical sectors \cite{TPAMI}. Large volumes of long-form pharmaceutical and medical videos, including clinical trial interviews \cite{tryout}, drug manufacturing workflows \cite{cellnet}, educational seminars, and long conference recordings, are routinely produced and remain difficult to consume, review, and reuse on scale.

Traditionally, the review and repurposing of such long-form videos is a labor-intensive process, often requiring days or weeks of manual effort by multiple stakeholders \cite{wu2018beyond, corinall, videocomsum}. Users typically rely on titles and brief summaries before manually navigating long recordings to assess relevance, a workflow that frequently leads to reduced interest. Generating concise highlight clips tailored to medical and pharmaceutical content can substantially reduce review time \cite{liu2020ultrasoundRL}, foster viewer engagement, and improve reuse of existing video assets \cite{liu2020ultrasoundRL, susarla2024youtubeChronic}. Recent research and commercial systems for video summarization and clip generation can be broadly categorized into three paradigms: \textbf{Frames-to-Video}, \textbf{Direct Video-to-Video}, and \textbf{Prompt- or Image-conditioned Video Generation}. 

\textbf{Frames-to-Video} approaches extract and assemble key frames into short clips, offering high customization allowing frame adjustments, removal, additions, along with audio manipulation, but incurring very long preprocessing and generation times, and often suffering from temporal discontinuities (e.g. jumping frames) \cite{opusclip2023ai, pikalabs2024frames, synthesia2024ai, heygen2024avatars}. \textbf{Direct Video-to-Video } methods process entire videos end-to-end, enabling faster generation and smoother outputs (less than 1-2 minutes for one short clip), but typically operate as black-box systems with limited transparency or control over clip selection, and post-selection requires more time to choose the best short clips.  \cite{wang2018video2video, liu2021uvit, bansal2018recyclegan, Li_2025_CVPR}.

%For \textbf{Frames-to-Video} methods, companies such as Oclips \cite{opusclip2023ai}, Pika \cite{pikalabs2024frames}, Synthesia \cite{synthesia2024ai}, or HeyGen \cite{heygen2024avatars} typically extract key frames to form a video. Characteristics of these methods include: (1) Video generation is slow . (2) Pre-processing is lengthy but accurate, requiring timestamps and summaries. (3) Consistency is vital, as jumping frames can occur. (4) High customization,  and highly realistic.

%\textbf{Direct Video-to-Video } methods \cite{wang2018video2video, liu2021uvit, bansal2018recyclegan, Li_2025_CVPR} features typically include: (1) Video generation incurs shorter times , offers high accuracy and smooth output without issues of jumping frames, and requires the original video file (API, YouTube, or cloud link). (2) Lack of customization, often acting as an enterprise "black box" for frame selection, and post-selection requires more time to choose the best short clips. 

Recent advances in VLMs and ALMs have enabled prompt-based video generation and multimodal understanding \cite{sora2024geometry, yang2025vlm_planning}. Companies like Stability AI \cite{stabilityai2024stablevideo}, Sora \cite{sora2024geometry}, Elevenlabs \cite{elevenlabs2024videonarratives}, DeepBrain \cite{deepbrain2024deepbrain}, Kaiber \cite{kaiber2024image2video}, and Animoto \cite{animoto2024animoto} adopt \textbf{Single Image+Prompts-to-Video } methods, which can produce visually coherent clips efficiently, but limited in Image/Video duration length (e.g single-image input, enterprise APIs may access frames, <20MB, max approx. 1 hour (video only) input) and computationally expensive. Their direct application remains ill-suited for clinical or pharmaceutical settings where preserving original frames (e.g., expert explanations, interviews, or procedural steps), exact source timestamps \& original audio, and traceability are critical \cite{hu2024ophclip, jiang2025omniv_med}. This setting introduces additional constraints, including long video durations (<2 minutes to 3 hours), strict latency and cost requirements, and the need for auditable, role-specific clip selection.

In this paper we investigate the following center research questions:
\textbf{RQ1:} Can an hybrid ALM/VLM-based pipeline generate high-quality highlight clips from long medical videos under strict non-synthetic, efficiency and cost constraints?
\textbf{RQ2:} How to improve temporal coherence and transition smoothness compared to direct video flame concatenation?
\textbf{RQ3:} How do role-based prompt personas influence clip selection behavior?

%In this paper,  we maintain high-resolution keyframe sampling and preserve original audio/text alignment, prioritizing clinical accuracy over purely synthetic animation.  We introduce a novel \emph{Infinite Video‑to‑Video Clips Generation} framework that leverages VLMs to facilitate the continuous extraction and generation of clinically-relevant video segments. These segments will be specifically tailored for pharmacyceutical content consumption, regulatory review on industrial multimodal GenAI platform. Our contributions include:

To address these questions, we propose an \emph{Infinite Video-to-Video Clips Generation} framework designed for long-form pharmaceutical and medical videos. Our contributions are as follows:
%Unlike prior generic summarization pipelines, our approach emphasizes (a) \textbf{traceability and frame preservation}, avoiding synthetic content that could compromise clinical accuracy; (b) \textbf{transparency and personalization}, allowing users to define roles and segment priorities; and (c) \textbf{efficiency and cost-awareness}, critical for large-scale VLMs industrial deployment. 

\begin{figure*}[ht]
  \centering
  % first row of three
  \begin{subfigure}[t]{0.32\textwidth}
    \centering
    \includegraphics[width=\linewidth]{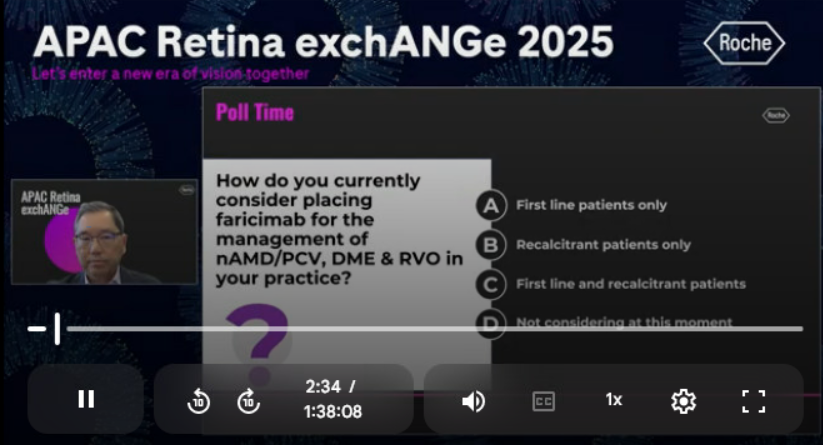}
    \caption{Original Long Seminar Video (1.5h)}   % optional
  \end{subfigure}\hfill
  \begin{subfigure}[t]{0.32\textwidth}
    \centering
    \includegraphics[width=\linewidth]{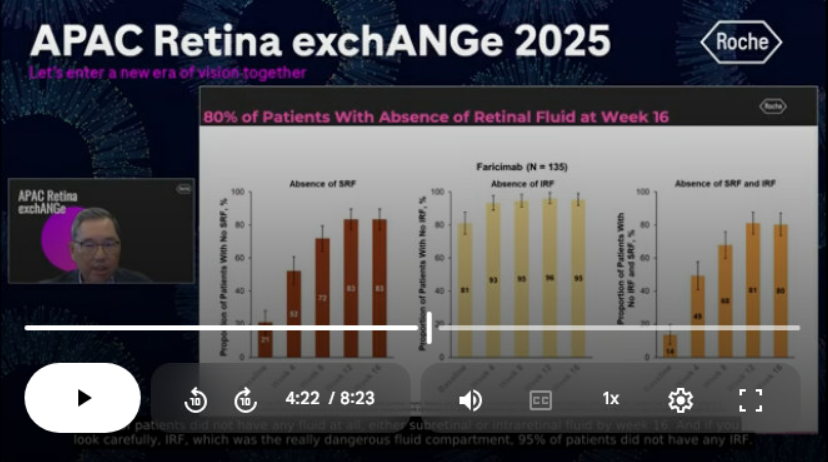}
    \caption{Ours w/ defined Output Length }
  \end{subfigure}\hfill
  \begin{subfigure}[t]{0.32\textwidth}
    \centering
    \includegraphics[width=\linewidth]{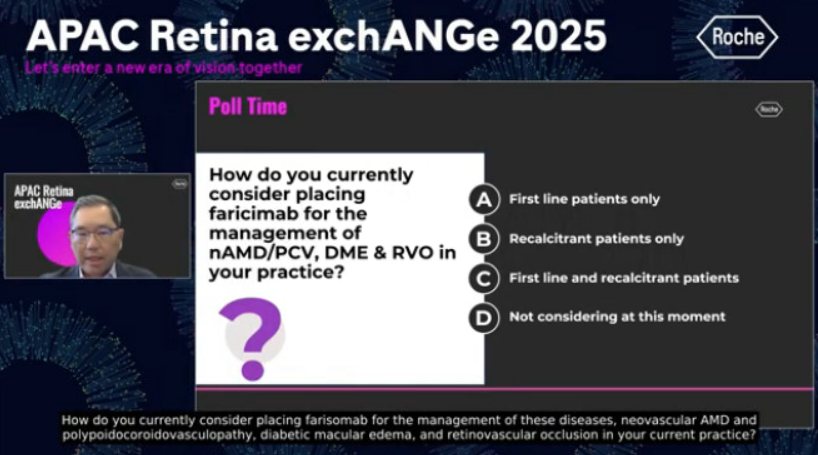}
    \caption{Ours w/ Subtitles and covering Agenda }
  \end{subfigure}

  \vspace{1ex} % small gap between rows

  % second row of three
  \begin{subfigure}[t]{0.32\textwidth}
    \centering
    \includegraphics[width=\linewidth]{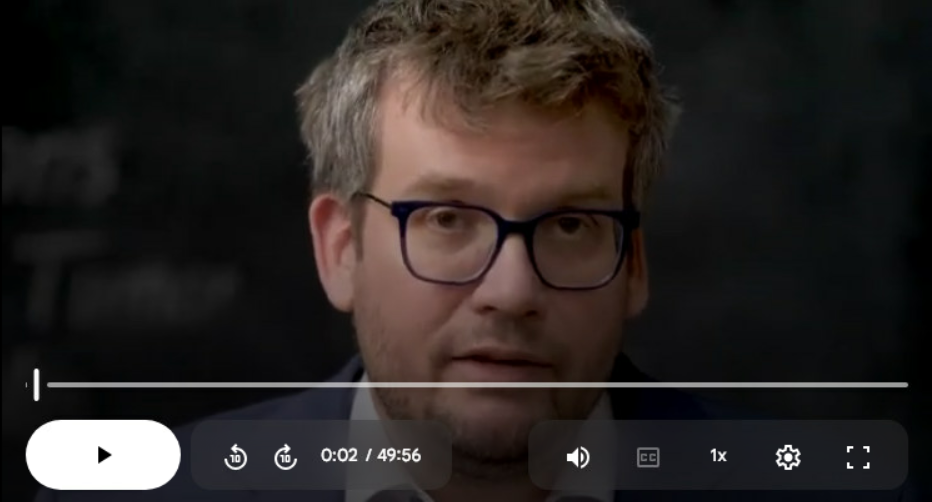}
    \caption{Original Long Lecture Video (0.8h)}
  \end{subfigure}\hfill
  \begin{subfigure}[t]{0.32\textwidth}
    \centering
    \includegraphics[width=\linewidth]{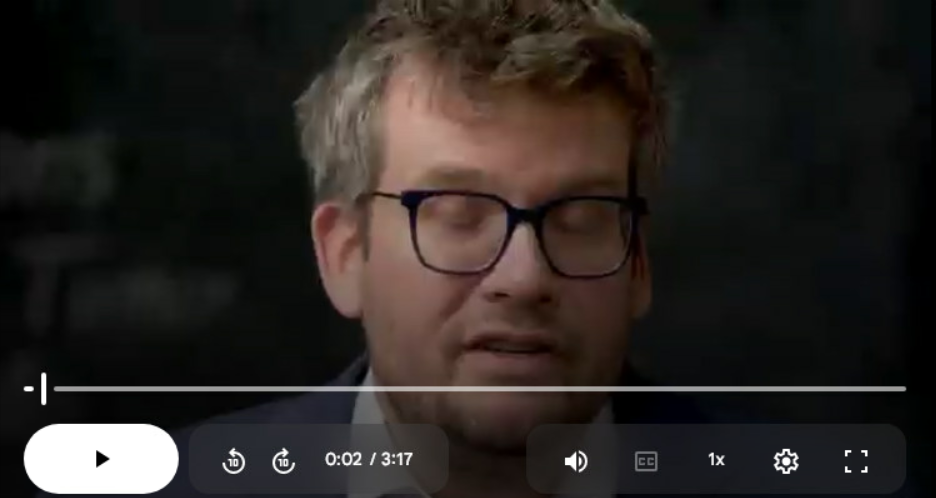}
    \caption{Frames to Video methods}
  \end{subfigure}\hfill
  \begin{subfigure}[t]{0.32\textwidth}
    \centering
    \includegraphics[width=\linewidth]{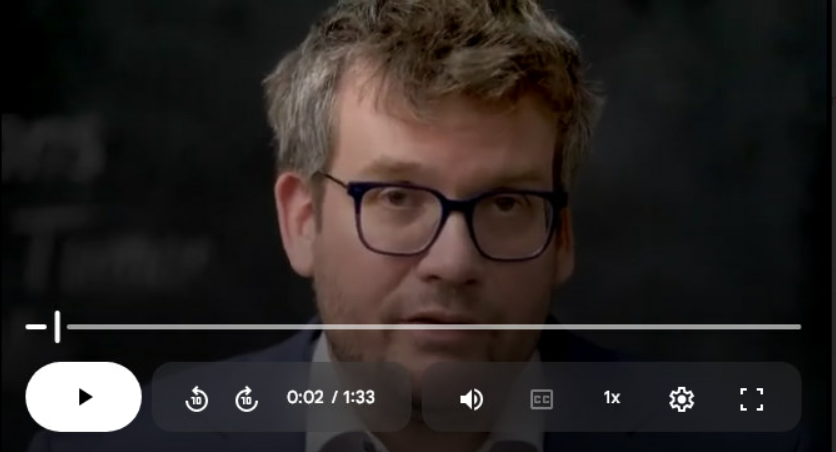}
    \caption{Ours}
  \end{subfigure}

  \caption{%
   \textbf{Qualitative comparison of our Infinite Video-to-Video Clips pipeline against frame-based approaches} (e.g.\ Runway Gen-2 \cite{runwayml2023gen2}). Our method supports arbitrary input durations, allows user-defined output lengths, automatically extracts agenda-relevant segments, adds subtitles and vertical playback, while overcoming choppy transitions and frame skipping/freezing, e.g, (\textbf{e}).
  }
  \label{fig:BenchmarkData}
\end{figure*}

\begin{itemize}
\item Cut \& Merge Algorithm: A reproducible, patent-pending algorithm that normalizes timestamps and applies fade-in/out boundaries to eliminate jump cuts, audio glitches, and frame freezing (see Fig:\ref{fig:BenchmarkData}, Alg:\ref{alg:clip_pipeline})
\item Personalization Mechanisms via Role Definition \& Prompt Injection: Systematically tailoring clip generation (e.g., promotional vs. educational styles) while maintaining transparency (see Fig:\ref{fig:personalization}, \ref{fig:demo}, \ref{tab: thinkingchain}).
\item Infinite Video-to-Video Clips Generation framework, as illustrated in Fig:\ref{fig: architecture}, suitable for industrial production adoption, characterized by lower cost 4x, 3-4x high speed  (see Table \ref{tab: mainresults}, Figure \ref{fig: finding 1}) and better clip-quality scores (Clip Coherence, Informativeness, Redundancy scores in Table \ref{tab:coherence}).
\item We also present five key technical findings substantiated by comprehensive evaluation, namely, speech detection and voiceover extraction, a security safeguard for the Whisper model, multimodality versus single modality, alignment and clip personalization.
\end{itemize}

%These instructions are for authors submitting papers to *ACL conferences using \LaTeX. They are not self-contained. All authors must follow the general instructions for *ACL proceedings,\footnote{\url{http://acl-org.github.io/ACLPUB/formatting.html}} and this document contains additional instructions for the \LaTeX{} style files.

\section{Related Work}

Research on video summarization and generation has evolved rapidly in recent years. Diffusion-based approaches \cite{ho2022video, wang2025survey, xing2023survey} have become the dominant paradigm video generation, replacing earlier GAN- and autoregressive-based methods \cite{vondrick2016generating, yan2021videogpt}. While diffusion models achieve impressive visual fidelity, scaling them to long videos remains challenging due to high computational costs, poor frame coherence and consistency. For long video processing — essential in domains such as medical procedure, patient interview analysis, and pharmaceutical education — recent work has focused on compressing or summarizing video content into token representations suitable for VLM input. Specific works like Video‑XL \cite{shu2024video_xl}, FiLA‑Video \cite{guo2025fila_video}, and LongVLM \cite{weng2024longvlm} have shown promise by using token-compression architectures to balance global context and local detail. 

The rise of vision-language models (VLMs) has further enabled multimodal understanding across video, image, and audio modalities \cite{sora2024geometry, yang2025vlm_planning}. Benchmarks such as Video-MME \citep{fu2024video} provide standardized evaluation protocols for long-video comprehension. Large-scale models including Qwen-VL \citep{Qwen-VL}, Gemini 2.5 \citep{deepmind2025gemini25}, and GPT-4o \citep{openai2023gpt4} demonstrate strong performance on alignment and summarization tasks. However, these systems are often costly to deploy at scale, operate as black-box models with limited transparency, and typically focus on video-to-text outputs rather than extractive video-to-video generation, particularly for long-duration inputs.
Several recent works explore LLM-driven video summarization and clip generation. Lee et al.~\citep{lee2025videosumm} demonstrate the potential of LLMs for clip generation but highlight challenges in maintaining logical flow and informativeness, e.g. generate synthetic or poorly aligned content that risks clinical misinterpretation \cite{guo2024youtube}, \cite{tariq2025vlmhealthcare} or fail to provide customize clips.
Representative academic pipelines such as \textit{LLMVS} use frame captioning followed by LLM-based scoring to generate summaries optimized for standard video summarization benchmarks \cite{lee2025video}. \textit{V2Xum-LLM} explores cross-modal video-to-text and video-to-video summarization using instruction-tuned datasets such as Instruct-V2Xum \cite{hua2024v2xum}. Prompt-to-Summaries methods enable zero-shot or query-controlled video skimming, but are primarily designed for short, publicly available video \cite{alaa2024videosummarizationtechniquescomprehensive}. 

While these approaches demonstrate strong performance on academic benchmarks, they operate under assumptions that differ fundamentally from our setting. In pharmaceutical video-to-video clip generation, the primary objective is not to synthesize visually appealing transitions or animations, but to identify and extract clinically meaningful segments (e.g., interviewer speech, key frames, or specific clinical interventions). Preserving original frames and audio is often more critical than generating new, visually enhanced footage, particularly under compliance and traceability requirements \citep{hu2024ophclip, jiang2025omniv_med}. Moreover, real-world industrial deployments must handle videos ranging from under two minutes to over three hours, mostly long videos, significantly exceeding the duration of standard benchmark datasets. Such systems often face strict latency, cost, and privacy constraints across multiple commercial ALM/VLM backends, while ensuring that extracted clips remain auditable and attributable to exact source timestamps.
These constraints limit the direct applicability of video generation or short video-to-text summarization pipelines, and motivate an extractive, hybrid ALM/VLM design that balances scalability with clinical reliability, a key requirement for deploying VLM-powered video-to-video systems in the pharmaceutical industry.

\section{Dataset And Experimental Settings}
%\subsection{Dataset Simulation}
\label{ssec:subhead}
\begin{table}[ht]
    \centering
    \caption{Statistics of Our Proprietary  Dataset.}
    \resizebox{0.50\textwidth}{!}{  % Resize the table to fit the page width
    \begin{tabular}{ll}
        \toprule
        \textbf{Category} & \textbf{Details} \\
        \midrule
        VLM Models Covered   &  Gemini 1.5 Pro. 2.0-Flash, Gemini 2.5 Pro \\ & Gemini 2.5 Flash,  Qwen-7B-VL, Qwen-72B\\ & Claude 3.5 Sonnet,  GPT-4o.\\

        ALM Models Covered  & Whisper-turbo V3 and Whisper-large V2. \\
        Number of Videos    & Over \textbf{16,159} Long Videos. Sampled 300 \\
        Number of Audios     & Over \textbf{888}. \\
        Covered Variants     & Over \textbf{14} Diseases areas. From Nephrology,\\ &   Ophthalmology to Hematology, Immunology, \\ &  Dermatology. \\
        Video  format Types     & \textbf{8 types:}. MP4, M4V, QuickTime, WMV, WebM, \\ &  MSVideo, MPG, and 3GPP. \\
        Audio format Types     & \textbf{4 types:}. '.mp3', '.wav', '.m4a', '.flac' . \\
        Video Lengths      & Major Longer length video from >30 min  \\ & to over 3 hours.\\
        Language Types       & Over 20 languages, including German, \\ &  Italian, English, Mandarin \\&Hokkien, Hindi, Korean, \\ &  French, Dutch, Spanish, and more. \\   
        \bottomrule
    \end{tabular}
    } \label{tab: RICI data}
\end{table}

Here, we primarily adopt well-established Video-MME \citep{fu2024video}, along with evaluations of major 11 SOTA VLMs, as well as our pharmacycy proprietary  dataset shown in Table~\ref{tab: RICI data}, Figure ~\ref{fig: BenchmarkData}.

Video-MME \citep{fu2024video} is the first full-spectrum multi-modal evaluation benchmark designed specifically for video-based MLLMs. It stands out from existing benchmarks with several key features: (1) Diversity in video types, covering six primary visual domains with 30 subfields to ensure broad scenario generalizability; (2) Temporal coverage, including short-, medium-, and long-term videos ranging from 11 seconds to 1 hour. It comprises 900 manually selected videos, totaling 254 hours, with 300 videos in each categories.

Furthermore, our findings are validated using proprietary data from 14 disease areas, including sampled over 300 long-form videos (<2mins minutes to over 3 hours, 8+ formats) and 888 audio.

\section{Methodology, Business Impact And Technical Features}

As illustrated in Figure\ref{fig: architecture}, our pipeline integrates ALMs, VLMs, various practical libraries, prompt-based segment selection, and a Cut \& Merge post-processing algorithm to generate extractive highlight clips from long-form videos.

\begin{figure*}[ht]
  \centering
  % first row of three
  \begin{subfigure}[t]{0.5\textwidth}
    \centering
    \includegraphics[width=\linewidth]{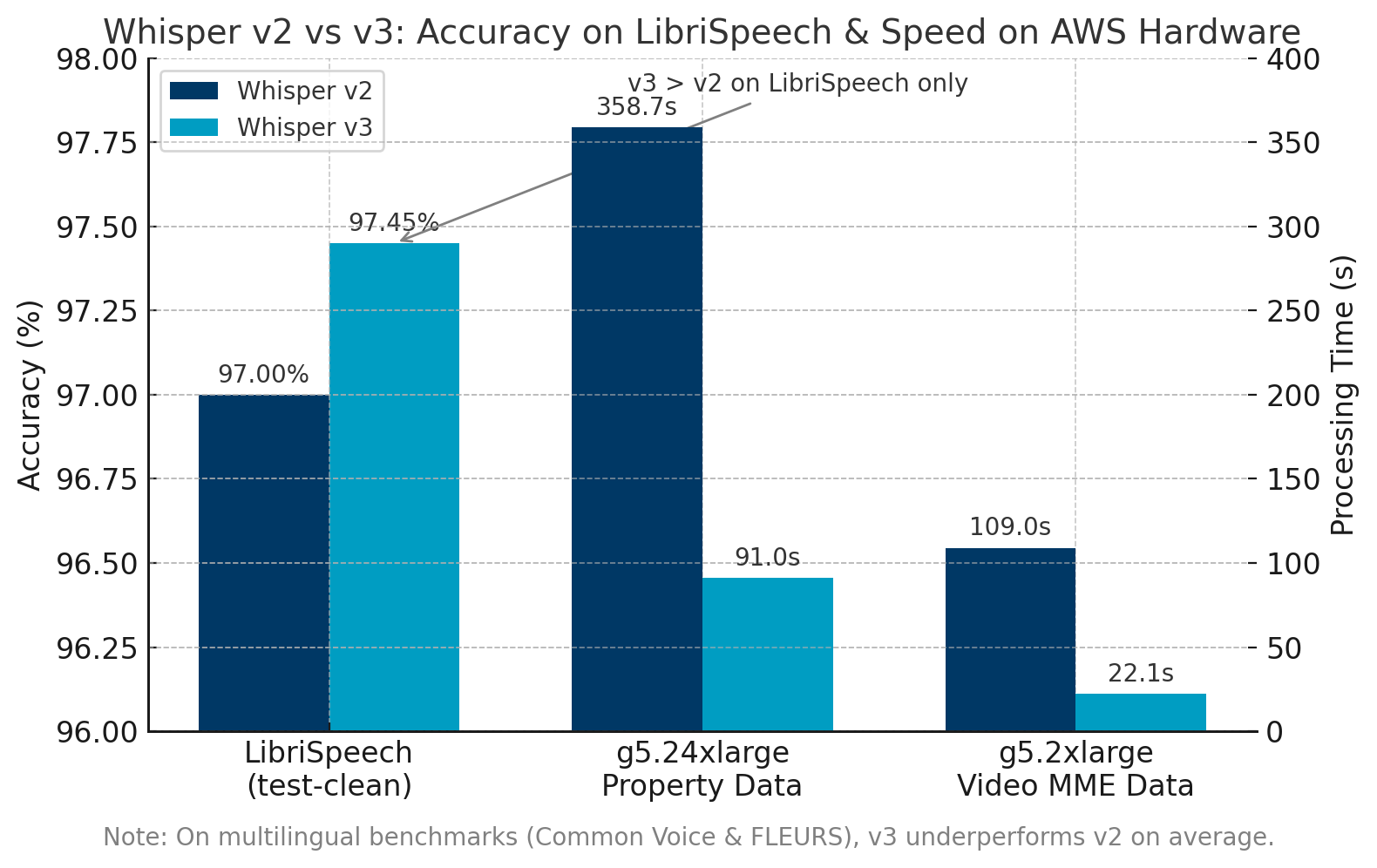}
    \caption{Accuracy On LibriSpeech \& Speed On AWS Hardware.}   % optional
  \end{subfigure}\hfill \hfill
  \begin{subfigure}[t]{0.5\textwidth}
    \centering
    \includegraphics[width=1\linewidth]{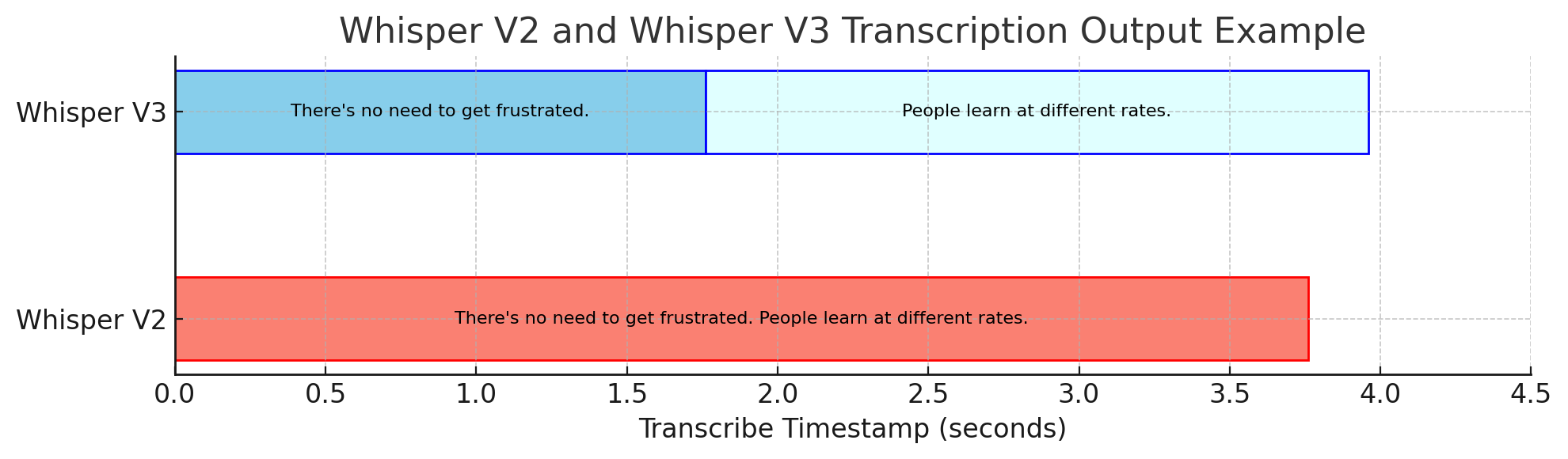}
    \caption{V3 has more fragment transcriptions (upper V3).}
  \end{subfigure}\hfill \hfill
  
  \caption{%
  \textbf{Qualitative comparison of Whisper V2 vs. V3: transcription accuracy on LibriSpeech \citep{panayotov2015librispeech} test-clean and inference speed on AWS hardware}. Whisper V3’s performance is more sensitive to GPU type—achieving roughly 4–8× speedups and higher Accuracy compared to V2, but cuts speech more aggressively, resulting in increased sentence fragmentation that complicates downstream timestamp alignment and segment merging. 
  }
  \label{fig:v3v2}
\end{figure*}

Given an input video, we first extract voiceover transcriptions using Whisper V2 and V3 ALMs \citep{radford2022whisper, openai2023whisper_v3} by a predefined schema. We employ both versions for quality control: while Whisper V3 provides four times faster processing and more accurate language type detection, Whisper V2 often yields more complete sentence boundaries. This complementary behavior is analyzed in Table~\ref{tab: mainresults} and Figure~\ref{fig:v3v2}. The resulting transcripts are then aligned with precise timestamps and serve as one of primary textual input for downstream segment selection.

Using the aligned transcription, we extract candidate video segments through structured prompt injection under four constraints:(1) user-defined target clip length, (2) video role or style (e.g., marketing, training, educational etc.), (3) audio--visual alignment with smooth fade-in and fade-out transitions, and
(4) Segment selection metrics. These metrics prioritize full-video coverage (including beginning and end segments), workflow transitions, agenda-highlighted content or video introduction, and audio cues such as pauses or changes in speaker intonation. For videos without voiceover, VLM-based visual understanding is used to identify salient segments based on visual content alone.

Rather than directly concatenating selected segments using e.g. FFmpeg, we designed Cut \& Merge algorithm (Algorithm~\ref{alg:clip_pipeline}) to ensure visual\& audio coherence and smooth transitions. Given the precise segment timestamps produced by our ALM/VLM pipeline, each segment is then processed individually as follows: (i) audio fade-in and fade-out are applied within a fixed temporal window to suppress background noise and abrupt audio transitions; (ii) video frames are re-encoded with synchronized visual fade-in/out (typically within a $\pm$0.5\,s window) to mitigate frame freezing and incompatibilities issues; (iii) processed segments are then concatenated in their desired order to produce the final clip, where user can also enlarge or remove segments slices (optional). Figure~\ref{fig:BenchmarkData} (e-f) compares our method with naive concatenation using standard tools (e.g., FFmpeg), illustrating improved transition smoothness and reduced skipping/freezing frames.

%Using these transcription files, we then extract key segments guided by four constraints via Prompt Injections: (1) User-defined output video length, (2) Video type/style (e.g., marketing, training, promotional, educational), (3) Audio alignment with fade-in and fade-out transitions, and (4) Segment selection metrics. These metrics include full-length coverage (especially ensuring inclusion of the beginning and end), emphasis on workflow and transitions, key information presented in the agenda or video introduction, and highlights such as speaker pauses or intonation peaks in audio. For video does not have voiceover we still have VLMs' feature to watch all virtual info through the videos. 

%Finally, instead of directly concatenating all segments, we then apply our Cut\&Merge Algorithm \ref{alg:clip_pipeline} to produce the final video clip. comparing to naive concatenation using e.g. FFMPEG, we first got precise segments from timestamp indicates from VLM/ALM pipeline, and then we for each segments we then do fade in and out in audio by given time windows to remove the voice noise in background, then for visual frame we do encoding, and also +- 0.5 time windows fade in and out to remove the frame freezing or compatibility issues. In the end we run the concatenation of all those processed segments and merge them according by given index of segments. we showed the visual difference our vs the naive concatenation in figure 2 e vs f.

\begin{algorithm}[t]
\caption{Cut \& Merge Clip  (Fade In/Out + Re-encoding)}
\label{alg:clip_pipeline}
\small
\begin{algorithmic}[1]
\Require $\mathcal{S}$ (selected segments with \texttt{start}, \texttt{end}), video $V$, output dir $D$
\Ensure merged highlight clip $V_{\text{out}}$
\State \textbf{mkdir}$(D)$
\State $\mathcal{L} \gets [\ ]$ \Comment{concat segment list file entries}
\For{$i \gets 1$ \textbf{to} $|\mathcal{S}|$} 
    \State $(s,e) \gets (\mathcal{S}[i].\texttt{start},\; \mathcal{S}[i].\texttt{end})$ \Comment{Timestamp}
    \State $p \gets D/\texttt{clip\_}i\texttt{.mp4}$
    \State \Call{ProcessClip}{$V, s, e, p$} \Comment{w/ fade + re-encode}
    \State $\mathcal{L} \gets \mathcal{L} \cup \{\texttt{"Processed segment '}\,p\,\texttt{'" }\}$
\EndFor
\State \Call{ConcatClips}{$\mathcal{L}, V_{\text{out}}$} \Comment{ffmpeg concat demuxer}
\end{algorithmic}
\vspace{0.3em}
\hrule
\vspace{0.3em}
\footnotesize
\begin{tabular}{@{}p{\linewidth} }
\textbf{ProcessClip}$(V,s,e,p)$ \\ 
\texttt{ffmpeg -y -ss s -to e -i V -vf format=yuv420p 
-af "afade=in:0:0.5,afade=out:(e-s-0.5):0.5"  \Comment{audio fade-in/out} 
-c:v libx264 -preset fast -crf 23 \Comment{Avoid direct concatenation `-c copy` } 
-c:a aac -b:a 128k p} \Comment{re-encoding to avoid jump cuts and frame freezing} \\
\\[0.3em]
\textbf{ConcatClips}$(\mathcal{L},V_{\text{out}})$ \\ 
\texttt{ffmpeg -y -f concat -safe 0 -i list.txt -c copy V\_out} 
\Comment{concatenate processed clips using FFmpeg concat demuxer}
\end{tabular}
\end{algorithm}

Our framework supports both vertical and horizontal playback formats and optional subtitle integration. It is designed to handle input videos of arbitrary length, overcoming the duration and storage constraints of many VLM-based systems (e.g. 20MB up to 1GiB). The pipeline offers transparent customization by allowing users to tailor video clips and remove or enhance specific sections. It is optimized for scalable deployment, achieving a 94.44\% time effort reduction for longer video summarization and 88\% time reduction across all pharmacy video categories compared to manual inspection. Finally, medical experts qualitatively assessed whether extracted clips preserved factual correctness and speaker intent. This evaluation is intended to assess alignment and usability in real life, and representative examples are reported in Appendix Table~\ref {tab:phesgo_appendix}.

%Our approach also enables vertical or horizontal playback formats and supports subtitle integration. Below, we summarize the key features of our method.
%(1) Infinite Video-to-Infinite Video:  we are capable of seamless integration between video sources of any length. (2) High transparency and customization: allowing users to tailor video clips and remove or enhance specific sections.  (3) Scalability and Cost Efficiency: designed for large-scale production,  We included medical experts verifying whether extracted clips preserved factual correctness and speaker intent. This evaluation is done qualitative and intended to assess alignment, not regulatory approval.(As shown sample results in Appendix Table 6.)

\begin{figure*}[ht]
    \centering
    \begin{minipage}{0.4\textwidth}
        \centering
        \includegraphics[width=1\linewidth]{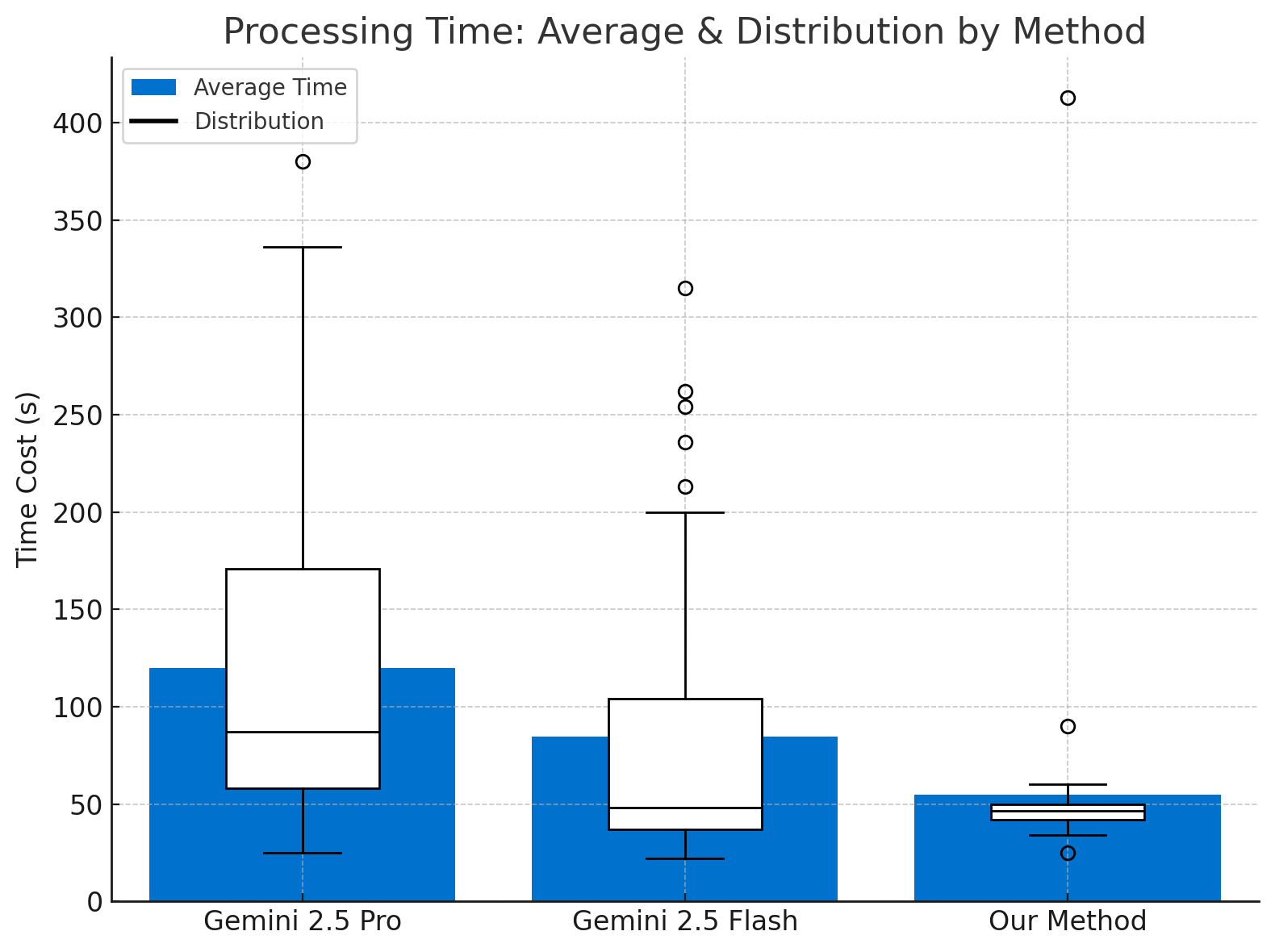}  
    \end{minipage}%
    \hfill
    \centering
    \begin{minipage}{0.6\textwidth}
        \centering
        \includegraphics[width=1.1\linewidth]{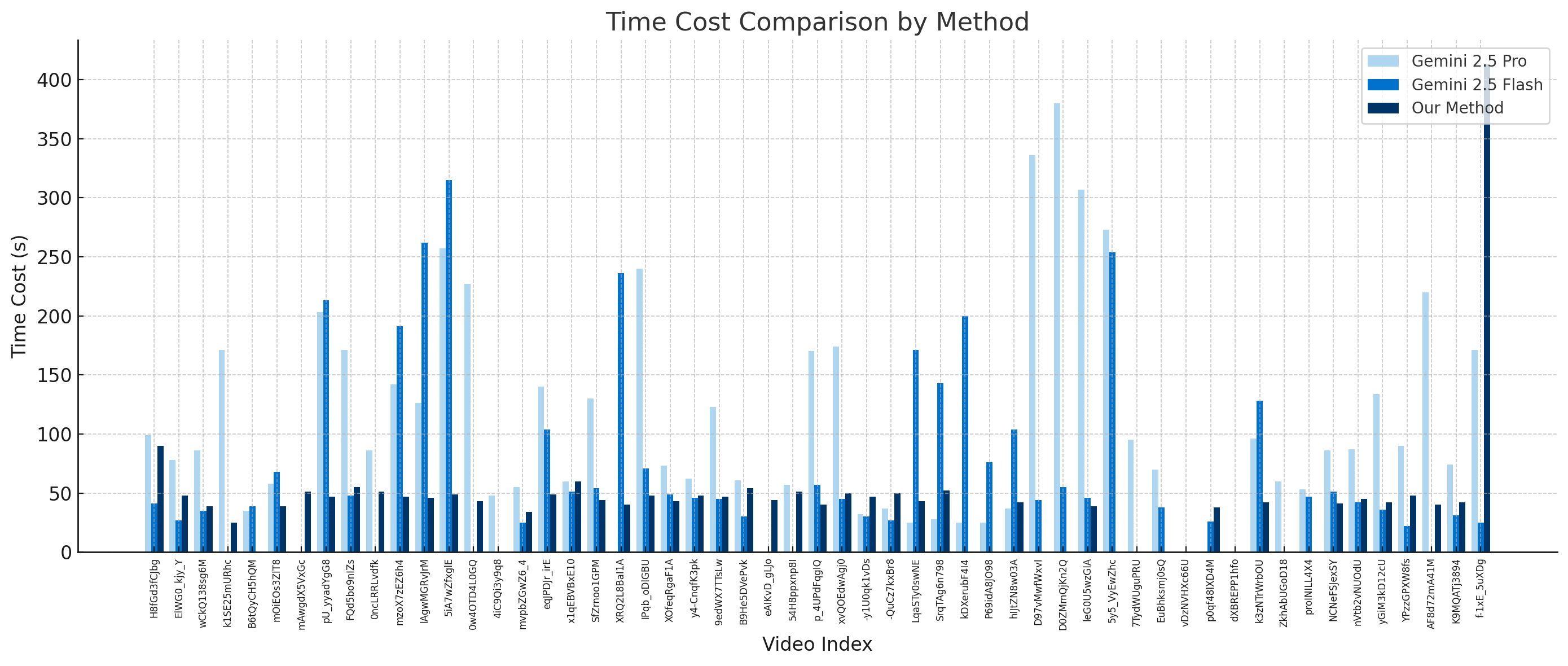}  
    \end{minipage}%
    \hfill
    \vspace{-10pt}
    \caption{ \textbf{Processing Time Comparison}: Gemini 2.5 Pro / Flash vs Our Methods for generating video clips script on VideoMME Long Video Dataset. Gemini Pro is the slowest (avg. $\sim$ 120s/video), with peaks on longer videos (e.g.,380s). Flash is faster ($\sim$80–85s) but still slower than our method ($\sim$30–55s), except for one outlier (413s). Our method is in general 3-4x faster.}
    \label{fig: finding 1}
\end{figure*}

\begin{figure*}[ht]
    \centering
    \begin{minipage}{0.5\textwidth}
        \centering
        \includegraphics[width=1\linewidth]{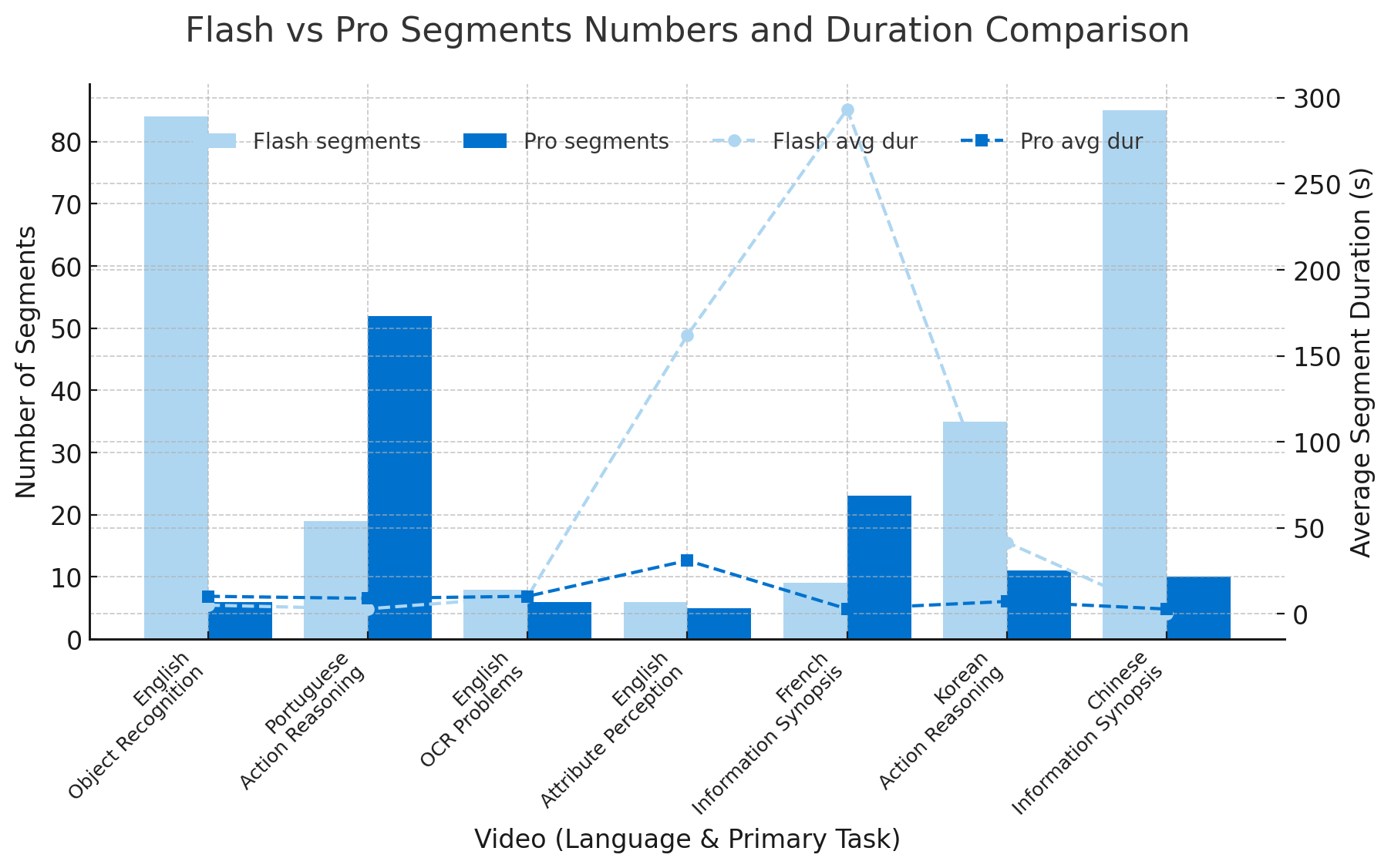}  
    \end{minipage}%
    \hfill
    \centering
    \begin{minipage}{0.5\textwidth}
        \centering
        \includegraphics[width=1\linewidth]{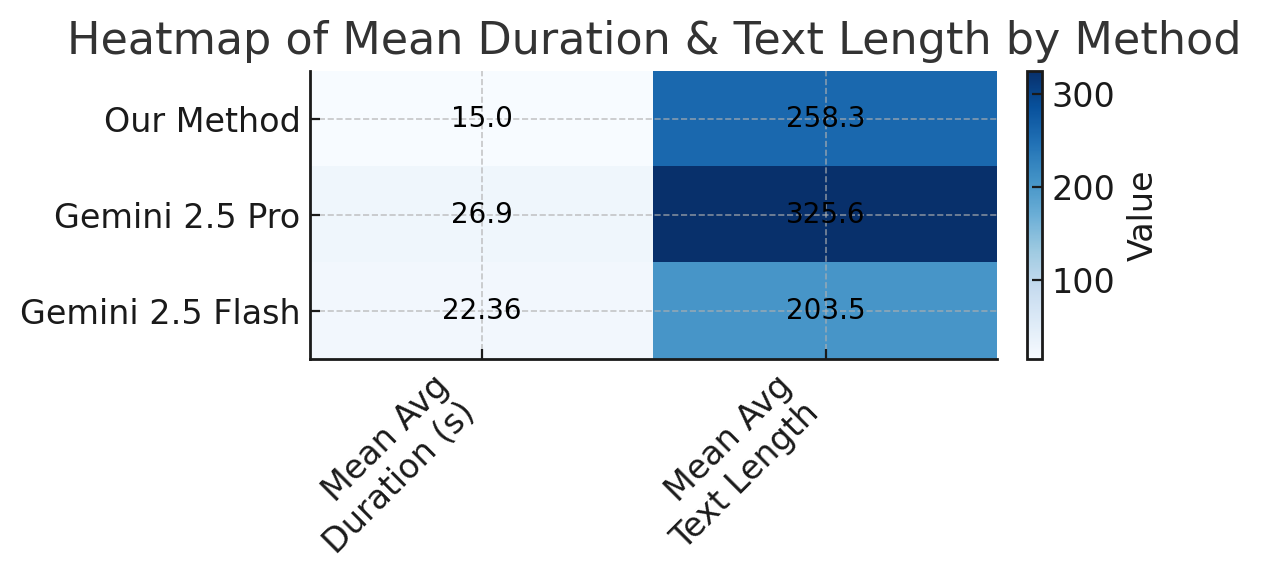}  
    \end{minipage}%
    \hfill
    \vspace{-10pt}
    \caption{\textbf{Comparison between the Gemini 2.5 Flash, Pro vs our, based on the number of select segments and the quality.} Here, we assess quality based on factors like segment length / numbers or the presence of coherent text. Flash often returns many but fragment segments (e.g. “Video ID: tslKtm6Le1s”: 85 piece of segments). Pro tends to pick fewer, longer segments (reflected in its lower segment counts but higher average durations). Our method achieves balanced selection 4.37 segments vs. 7.38 (Gemini 2.5 Pro) and 13.30 (Gemini 2.5 Flash) for final clips.}
    \label{fig: finding 2}
\end{figure*}

\begin{table*}[t]
\centering
\caption{\textbf{Processing time comparison in each step.} Experiment setting: Proprietary Dataset and VideoMME, SDPA, FPS=0.01, AWS ml.g5.24xlarge, g5.2xlarge instance. * means Gemini 2.5 Pro has Timeout for longer waiting or error cases. - means not applicable.}
\resizebox{\linewidth}{!}{%
\begin{tabular}{l|c|c|c|c|c|c|c}
\hline
\multicolumn{8}{|c|}{\cellcolor{Industry Proprietary  Datablue}\color{white}\textbf{Processing time cost (avg / per video, in seconds)}} \\
\hline
\rowcolor{tablehead}
\textbf{Video Type} & \textbf{VLM} & \textbf{ALM} & \textbf{FFmpeg/Moviepy} & \textbf{LLM} & \textbf{Cut \& Merge} & \textbf{Adding} & \textbf{Video amount} \\
\rowcolor{tablehead}
& (Directly Video to Text) & (Transcription by Whisper Turbo V3 & (Voice Over & (Bedrock Claude & \textbf{Video Clips} & \textbf{Subtitles} & \textbf{VideoMME} \\
\rowcolor{tablehead}
& & V3 vs Large-V2) & abstraction, & Sonnet /Gemini 2.5 Pro  & & & \textbf{ / Proprietary  Data}\\
\rowcolor{tablehead}
& & & video to audio & Extract Timestamp and & & & \\
\rowcolor{tablehead}
& & & file) & improve summary & & & \\
\rowcolor{tablehead}
& & & & quality) & & & \\
\hline
& Generate & \textbf{90.96s} (15.16 Mins in & & & & & \\
Longer video (& Summary & total, Turbo  v3 model) & & & & & \\
above 30 minutes& and Key & \textbf{358.74s} (59.79 Mins in & \multirow{-3}{*}{\textbf{79.2s}} & \multirow{-3}{*}{\textbf{24.37s}} & \multirow{-3}{*}{\textbf{30s-1Min}} & \multirow{-3}{*}{\textbf{4Mins}} & \multirow{-3}{*}{\textbf{300 / 100 }} \\
long)& Frame & total, Large-V2)  & & & & & \\
& Caption & (Proprietary  Data \& on g5.24xlarge) & & & & & \\
& 1-4 Min for & \textit{avg, 22.08s (3h for 489 videos on} & & & & & \\
& Gemini 2.5 & \textit{video mme by Turbo V3 model)  vs} & & & & & \\
& Pro * & \textit{avg, 109s (15h for 493 videos on } & & & & & \\
& & \textit{by Large-V2 model)}  & & & & & \\
& & \textit{(on g5.2xlarge)} & & & & & \\
\hline
& & \textbf{17.04s} (2.84 minutes & & & & & \\
Medium (2 minutes& \multirow{-2}{*}{\textbf{1-4 Mins}} & in total, Turbo V3) & \multirow{-2}{*}{\textbf{29.76s}} & \multirow{-2}{*}{\textbf{24.37s}} & \multirow{-2}{*}{\textbf{30s-1mins}} & \multirow{-2}{*}{\textbf{<4mins}} & \multirow{-2}{*}{\textbf{300 / 100}} \\
$<$ length $<$ 30& & \textbf{59.52s} (9.92 Mins in & & & & & \\
minutes)& & total, Large-V2) & & & & & \\

\hline
& & \textbf{4.26s} & & & & & \\
Short (less than 2& \multirow{-2}{*}{\textbf{-}} & (0.71 minutes & \multirow{-2}{*}{\textbf{10.56s}} & \multirow{-2}{*}{\textbf{-}} & \multirow{-2}{*}{\textbf{-}} & \multirow{-2}{*}{\textbf{-}} & \multirow{-2}{*}{\textbf{300 / 100}} \\
minutes)& & In total for turbo) & & & & & \\

\hline
\multicolumn{7}{c|}{\textbf{Total (avg)}}  & \textbf{1200} \\
\hline
\end{tabular}%
}\label{tab: mainresults}
\end{table*}

\section{Main Results}

\begin{table*}[t]
\centering
\small
\caption{\textbf{Overall Accuracy of VLMs in summarization on Video-MME.}
(A) Overall accuracy on Video-MME (900 videos) with/without audio transcription.
(B) Keyframe/time alignment and summary accuracy on long-video subset ($n=300$). Gemini models achieve high accuracy on meaningful summary, but still struggle with timestamps.}
\label{tab:vlm-overall-keyframe}

\begin{subtable}[t]{0.49\textwidth}
\centering
\subcaption{ Overall Acc. with/without audio transcription}
\begin{tabular}{l r r r}
\toprule
\textbf{Model} & \textbf{w/o } & \textbf{w/} & $\boldsymbol{\Delta}$ \\
\midrule
\textbf{Gemini 2.5 Pro}  & \textbf{84.7} & \textbf{85.2} & +0.5 \\
Gemini 1.5 Pro           & 75.0 & 81.3 & +6.3 \\
Qwen2\,-VL               & 71.2 & 77.8 & +6.6 \\
GPT\,-4o                 & 69.0 & 77.2 & \textbf{+8.2} \\
LLaVA\,-Video            & 76.0 & 76.9 & +0.9 \\
Gemini 1.5 Flash         & 72.6 & 75.0 & +2.4 \\
Oryx\,-1.5               & 67.3 & 74.9 & +7.6 \\
InternVL2.5              & 67.6 & 74.0 & +6.4 \\
Aria                     & 70.3 & 72.1 & +1.8 \\
LinVT                    & 65.6 & 71.7 & +6.1 \\
TPO                      & 66.2 & 71.5 & +5.3 \\
\bottomrule
\end{tabular}
\end{subtable}
\hfill
\begin{subtable}[t]{0.49\textwidth}
\centering
\subcaption{ Long-video Keyframe \& Summary Accuracy}
\begin{tabular}{l r }
\toprule
\textbf{Method} & \textbf{Summary Acc. (\%)} \\
\midrule
\textbf{Gemini 2.5 Flash} & \textbf{94.6} \\
Qwen\,-7b                  & 74.3         \\
\bottomrule

\textbf{Method}  & \textbf{KeyFrame Time Acc. (\%)} \\
\midrule
\textbf{Gemini 2.5 Flash}  & \textbf{35.1} \\
Qwen\,-7b                   & 5.4 \\
\bottomrule
\end{tabular}
\end{subtable}

\end{table*}

\begin{table*}[t]
\centering
\small
\caption{\textbf{Clip Coherence Scores / Informativeness based on Video-MME Benchmark (All Sampled Long Videos, $n=300$) @ $\tau=0.6$}. \textbf{Our Method:} Concise, 4$\times$ faster, 3.5--4$\times$ cheaper, consistently informative (0.721 informativeness, low redundancy 0.339). \textbf{Gemini 2.5 Pro:} Best logical flow but slower, more expensive, less consistent. \textbf{Gemini 2.5 Flash:} Overly long, fragmented, highly unpredictable summaries with high deviation.}
\label{tab:coherence}
\begin{tabular}{llllp{6cm}}
\toprule
\textbf{Metric} & \textbf{Our Method} & \textbf{Gemini 2.5 Pro} & \textbf{Gemini 2.5 Flash}  \\
\midrule
Clip Coherence scores (0-1) & 0.348 $\pm$ 0.118 & \textbf{0.446 $\pm$ 0.111} & 0.410 $\pm$ 0.127  \\
(How logically connected adjacent clips are) & & \\
\midrule
Informativeness scores (0-1) & \textbf{0.721 $\pm$ 0.078} & 0.674 $\pm$ 0.158 & 0.701 $\pm$ 0.124  \\
(How well summary clips represent the source) & & \\
\midrule
\# Segments number of clips & 6.46 $\pm$ 1.67 & 9.21 $\pm$ 10.40 & \textbf{18.32 $\pm$ 31.03}  \\

Redundancy Scores (Overlap among clips) (0-1) & \textbf{0.339 $\pm$ 0.106} & 0.415 $\pm$ 0.117 & 0.379 $\pm$ 0.138  \\

Clips meaningful text length & 258.3 & \textbf{325.6} & 203.5  \\

Speed(Generation time per clips) & \textbf{30--55s / video} & 120s avg / video & \textbf{80–85s avg / video}   \\

Cost in Dollar & \textbf{0.3 input / 2.5 output } & 1.25 input / 10 output  & 0.3 input / 2.5 output  \\
 & \textbf{ per M Token} & \textbf{ per M Token} & \textbf{ per M Token} \\
\bottomrule
\end{tabular}
\end{table*}

%In this section, we first evaluate with the top two benchmarks, Video-MME \citep{fu2024video} and MMBench \citep{MMBench} for multimodal vision language models (MVLMs), focusing on widely recognized leading vision language models (VLMs). Furthermore, we conducted over four iterations of experiments using our proprietary dataset, which includes the data presented in Table 1. 

%These experiments were designed to evaluate the performance of our findings in a generalized domain rather than in a specific domain to the medical industry.

%FFMPEG \cite{ffmpeg2024} capabilities extend to a vast array of multimedia tasks, including video and audio encoding, decoding, transcoding, muxing, demuxing, streaming, and filtering.

\textbf{How to secure the complex \& fast voice extraction?} In our pipeline, FFmpeg \cite{ffmpeg2024} is used for key operations such as format conversion and audio extraction to ensure broad compatibility. It general performs well on 1–2 hour videos with clean audio, averaging costs only 46.38 seconds per video. However, FFmpeg also encounter issues with long filenames, unsupported characters, incorrect encoding or compatibility errors. To handle more complex cases in reality, we also integrate MoviePy \cite{moviepy} as a fallback solution. MoviePy offers a Pythonic interface for programmatic and efficient audio extraction across video batches. While slightly slower (averaging 64.71 seconds), it delivers comparable extraction quality. This combined approach ensures robust audio processing across a wider range of video files/kinds, with extraction times typically ranging from 10 to 79 seconds, as shown in Table \ref{tab: mainresults}).

\textbf{Fragmented Sentences? Safe and accurate combination of Whisper V3/V2.} As illustrated in Figure \ref{fig: architecture}, \ref{fig:v3v2}), Whisper V3 \citep{openai2023whisper_v3} Whisper V3 offers significant speed improvements (up to 4x faster on industry datasets and VideoMME) and reliable language type detection. However, it often produces fragmented sentences, leading to unstable segmentation and less precise timestamp alignment(see Fig \ref{fig:v3v2}). In contrast, Whisper V2 \citep{radford2022whisper} Whisper V2 provides more accurate timestamps, crucial for tasks like clip-cutting, and is therefore our primary model. While V2 occasionally excels at capturing full sentences, real-world tests with English videos revealed sporadic language misidentification (e.g., mistaking English for Welsh), resulting in corrupted transcriptions. To address this, we use V3 as a secondary validation layer for security guidance  (see Figure \ref{fig: architecture}).

\textbf{Modality: A Key to Cost Reduction, Speed, Clips coherence} As shown in Figures \ref{fig: finding 1} and \ref{fig: finding 2}, utilizing ALM for accurate transcription, combined with Gemini 2.5 Pro (text-to-text), significantly reduces costs (up to 4-4.5× less, Table \ref{tab:gemini_pricing}) and accelerates processing time (up to 4× faster, 30-55 seconds per video) compared to using Gemini 2.5 Pro on full video input (average 120s, max 350s per video). Our method maintains comparable accuracy in generating less redundant, high coherent, highly informative video clip scripts (258 meaningful words) compared to Gemini 2.5 Pro (325) and Gemini 2.5 Flash (203). 

Furthermore, as shown in Table \ref{tab:coherence}, Clips coherence scores (fraction of original segments covered by any summary clips segment), Informativeness (cosine similarity to its best-matching original segment, averaged across segments), and Redundancy scores (mean pairwise cosine similarity among  segments (off-diagonal)), our approach efficiently provides comparable coherence (0.348), higher informativeness (0.721), reduced redundancy (0.339), and increased stability.

Gemini 2.5 Flash while generally faster (averaging 80–85 seconds per video) than 2.5 Pro, struggles more with non-English, abstract, or non-verbal video content. Gemini Pro tends to select fewer but longer segments, often influenced by visual information, leading to lower segment counts, higher average durations and standard deviations, and even reduced informativeness scores. Flash, conversely, selects more fragment but shorter clips (see Fig \ref{fig: finding 2}).

\textbf{Audio and Speaker Alignment: Direct Concatenation Will Not Work!} As shown in Figures \ref{fig:BenchmarkData}(e), directly concatenating video segments using frame-level and ALM-generated timestamps often results in frame jumps and background noise. To address this, we developed an algorithm that combines segment encoding (for noise reduction) with audio/video fade-in and fade-out transitions (+/- 0.5 seconds for smooth transitions). This approach avoids the visual and auditory glitches typically seen with direct FFmpeg concatenation. Furthermore, we standardize ALM-generated timestamps to two decimal places to improve segment precision and overall clip smoothness. Figures \ref{fig:BenchmarkData}(e)–(f) clearly demonstrate the advantages of our method: where other approaches show at the same timestamp a frozen eye or losing frames, but our transitions consistent with the original video.

\textbf{Personalization: Prompt Injection, Role Definition, and Selection Metrics}. In prompting, users can select a role and specify a maximum duration for video clips (e.g., up to 3-4 minutes). Additionally, the current process guides Gemini 2.5 Pro to explain its reasoning for selecting specific segments (See Appendix Figure \ref{fig: thinkingchainoferrorcase}, Table\ref{tab: thinkingchain}. This enables Gemini 2.5 to focus on key factors, such as smoothly transitions (finishing speaker's sentence), important key segments presented agenda topics/slides, keywords, full-length coverage, speaker's voice and pauses, and noun emphasis, rather than being distracted by numerous unrelated elements.

\section{Ablation study on Role Definition and Prompt injections}
How does each segment's metrics play a different role in the final outcome? We investigate how our segment metrics shape the final outcome by conducting an ablation that isolates those factors: (i) Keywords, ii) Agenda, (iii) Speaker-voice tone. As shown in Appendix Figure \ref{fig:personalization}, 1) Removing tone consistency can lead to visual-voice mismatches,  2) Agenda helps ensure clips retain key information, paragraphs, and segments, 3) Incorporating keywords (e.g., nouns, medical terms) increases the likelihood of important information being captured in segments, 4) Length coverage can prevents abrupt, incomplete clips, jump cut and sudden Ending. To support reproducibility, we open-source all role-specific prompt instructions in the appendix Figure \ref{tab:prompt_template} and to complement the demo link, we also upload additional sample clips based on non-sensitive healthcare videos: \href{https://storage.googleapis.com/ricivideoclips/Clips/Clips_Sample.zip}{Sample Video Clips}.

\section{Conclusion}

In this work, we introduce a novel industrial, practical video-to-video clip highlights framework. This addresses the critical need for efficient video summarization and highly customizable video clip generation from long-form content. Our framework significantly facilitates content reuse in industrial settings by enabling users to generate diverse clips cost-effectively, reducing processing time from hours to mere minutes. Our key contributions include: (i) Cut \& Merge algorithm that ensures smooth transitions through fade-in/out boundaries and timestamp normalization; (ii) Personalization mechanism using role definition and prompt injection to generate clips tailored for regulatory, educational, or promotional contexts; and (iii) end-to-end framework that leverages ALM pipelines for efficiency and VLM-enhanced processing for visual-heavy content. Evaluations on Video-MME and a proprietary dataset of long-form videos across 14 disease areas demonstrate 3–4× speedups, 4× cost reduction, and competitive clip quality compared to state-of-the-art VLM baselines.

We further highlight following key findings: 1) Modality matters, using voice transcriptions as a single modality can achieve very fast, high-quality clips at a very low cost. 2) Models like Gemini 2.5 Pro/Flash often rely on external tools (e.g., Google Search or Data APIs) rather than processing frame by frame, which can cause losing focus and produce fragmented segments. 3) Combining sophisticated prompt Injections —including transitions, agenda topics, keywords, full-length coverage, speaker voice/pauses — substantially enhances the final quality of the generated clips, surpassing the results of direct concatenation.

%\textbf{Limitation:}
%Despite being the first to release such large industrial dataset on accident information, our data still have limitations. To more effectively test the temporal and spatial awareness capabilities of LLMs, we need to manually annotate more spatial and temporal data and ground truths. Expanding to other regions or cities would require additional approvals from governments or institutions, which could further enhance our dataset.
%\textbf{Future work:}
%To address these limitations, we will continually collect accident information from various cities. Additionally, we plan to exploring various other functionalities of LLMs beyond just hallucinations.

%\section{Acknowledgments}  
%We sincerely thank Samik Adhikary and Puneet Srivastava for their sponsorship support from Industry Proprietary  Data. We also appreciate the insightful discussions and technical assistance provided by Janina Kummerfeldt, Philipp Müller, Kathrin Schwan from Accenture, as well as Aditya Kumar from Involead.  

\newpage
\clearpage

\section{Limitations}

Our methodology focuses on applying Vision–Language Models (VLMs) to life sciences, rather than proposing entirely new model architectures. A main limitation of this study is to focus on justification of the Pharm-specific benefits. We provide a baseline comparison of more than eleven VLMs using both the Video-MME benchmark and our proprietary dataset. Future work should extend this line of research to other regulated domains, such as financial services and manufacturing, to further validate the generalizability of framework.

\section{Acknowledgments}  
We sincerely thank Samik Adhikary and Puneet Srivastava for their sponsorship support from Roche. We also appreciate the insightful discussions and technical assistance provided by Janina Kummerfeldt, and Kathrin Schwan from Accenture, Jennifer McGuire's business support from Roche.
This platform, RICI, would not have been possible without their contributions. We further extend our gratitude to the backend engineering teams supported RICI development, as well as to the healthcare professionals (HCPs), testers, and Roche Lab users whose consistent feedback brought our Video Clips use cases and enabled continuous improvement.
Through this work, we aim to highlight current limitations of leading VLMs and ALMs and to contribute practical solutions to the research community, and share industry lessons learned and valuable large-scale GenAI experiments in the pharmaceutical domain.

\bibliography{custom}
\clearpage
\appendix
\section{Appendix}
\label{sec:appendix}

\begin{figure}[htbp]
  \centering
  % first row of three
  \begin{subfigure}[t]{0.5\textwidth}
    \centering
    \includegraphics[width=\linewidth]{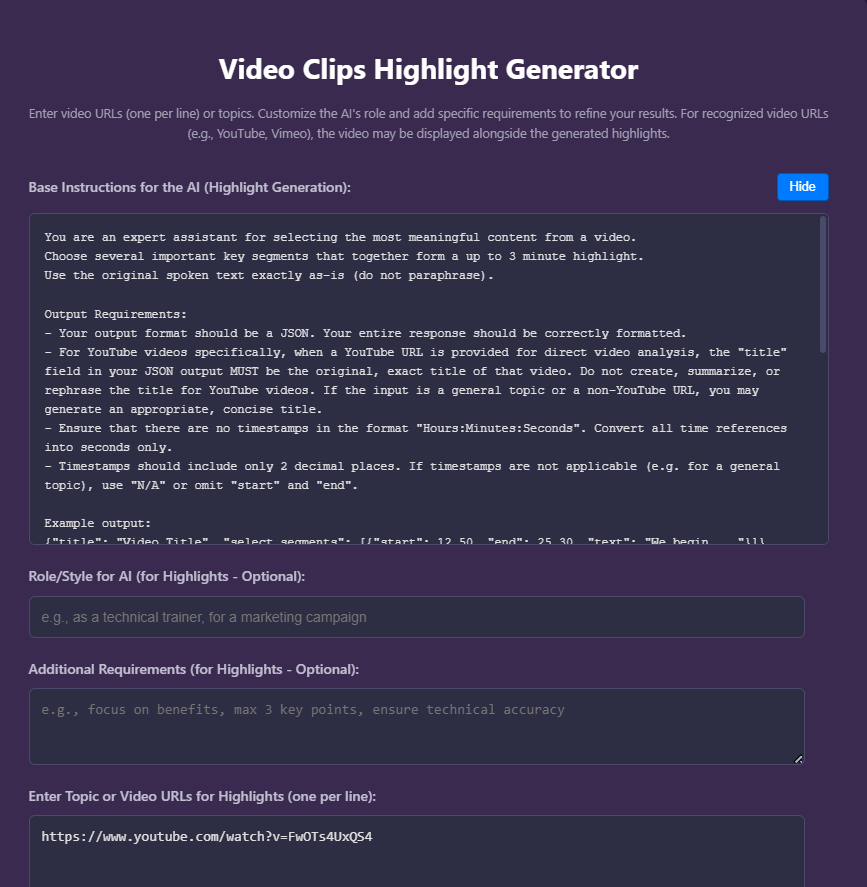}
    \caption{Prompt Logic}   % optional
  \end{subfigure}\hfill \hfill
  \begin{subfigure}[t]{0.5\textwidth}
    \centering
    \includegraphics[width=\linewidth]{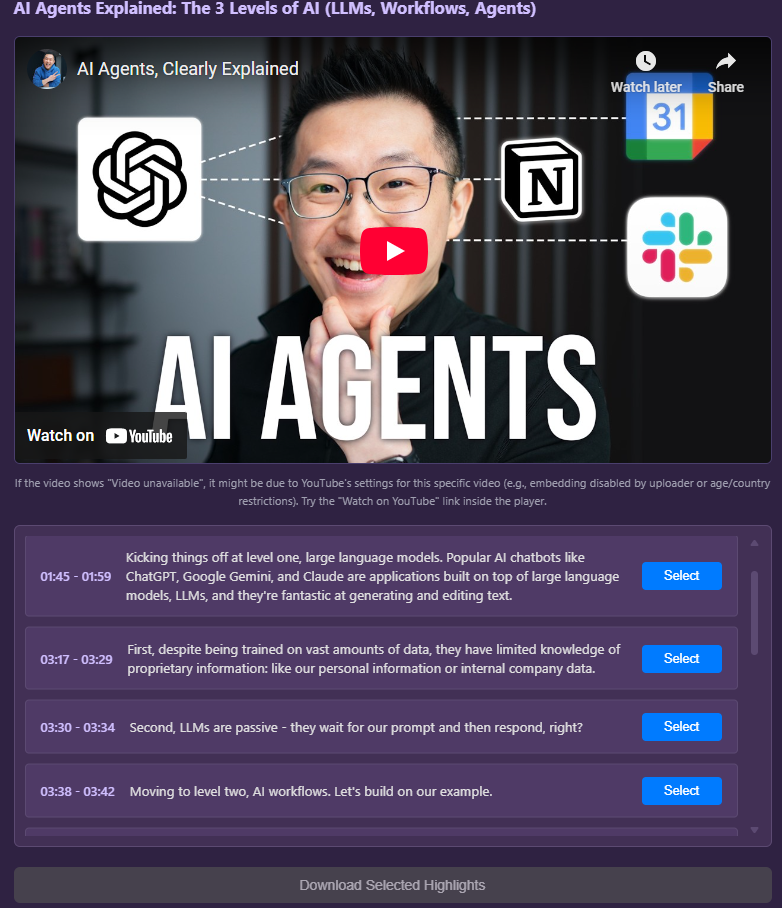}
    \caption{Generated Highlights Clips Scripts}
  \end{subfigure}\hfill \hfill
  
  \vspace{1ex} % small gap between rows
  \caption{%
   \textbf{Key Segments Selection Logic.} This logic has four main components: Role: Defining the role for GenAI in tasks. Output Requirements: Specifically on timestamp format for uniformity and error avoidance, and rephrasing needs. Example Output: Providing examples for clip script files. Core Task Definition: This includes segment selection criteria that cover key ideas, transitions, agenda points, distributed throughout the video (including beginning and end sections if relevant). For videos, consider the speaker's tone and pauses for smooth clip flow. Users can also input the role and additional requirements.
  }
  \label{fig:demo}
\end{figure}

\begin{figure*}[htbp]
  \centering
  % first row of three
  \begin{subfigure}[t]{0.32\textwidth}
    \centering
    \includegraphics[width=\linewidth]{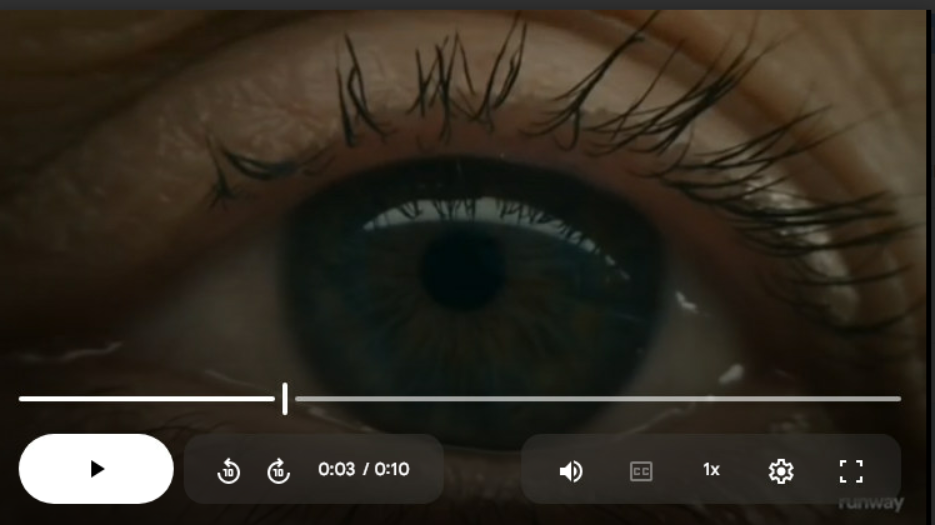}
    \caption{single-image-to-video (Runway)}   % optional
  \end{subfigure}\hfill \hfill
  \begin{subfigure}[t]{0.32\textwidth}
    \centering
    \includegraphics[width=\linewidth]{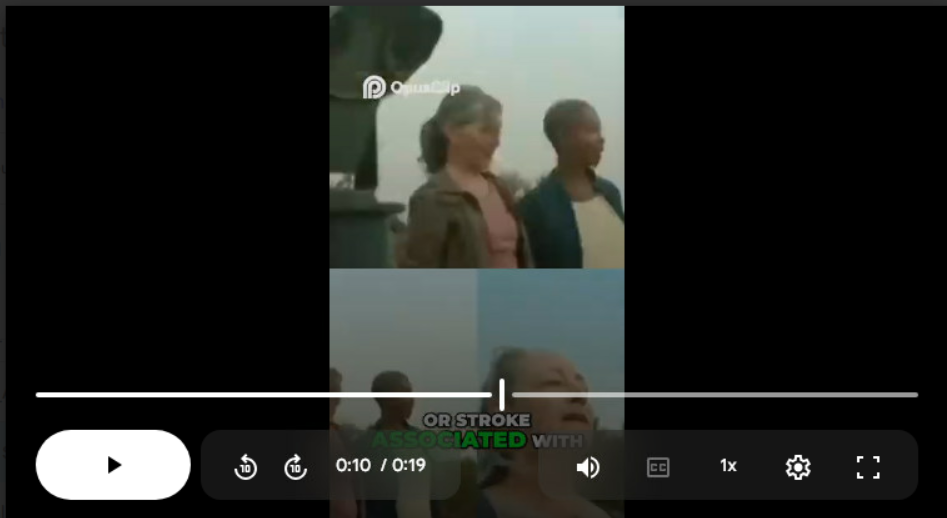}
    \caption{Frames-to-video (OpusClip)}
  \end{subfigure}\hfill \hfill
  \begin{subfigure}[t]{0.32\textwidth}
    \centering
    \includegraphics[width=\linewidth]{img/Ours_with_subtitles.png}
    \caption{Ours w/ Subtitles }
  \end{subfigure}

  \vspace{1ex} % small gap between rows
  \caption{%
   \textbf{Qualitative comparison of our Infinite Video-to-Infinite Video pipeline against single-image-to-video baselines and commercial tools} (e.g., Runway Gen-2 \citep{runwayml2023gen2}, OpusClip \citep{opusclip2023ai}). While existing methods generate from one or a few static frames, limited to <30 s outputs, prone to choppy transitions and frame skipping, and relying on shot-selection heuristics. Our methods support arbitrary input durations, user specified output lengths, optional subtitles, and vertical playback.
  }
  \label{fig:BenchmarkData 7}
\end{figure*}

\begin{figure*}[t]
  \centering
  % To ensure vertical alignment, we can add \vfill inside the subfigures
  % or use the [t] option as you did. The main fix is in \includegraphics.

  % ---------- Row 1 ----------
  \begin{subfigure}[t]{0.32\textwidth}
    \centering
    % Corrected: removed 'keepaspectratio' to enforce uniform size
    \includegraphics[height=3.2cm, width=\linewidth]{{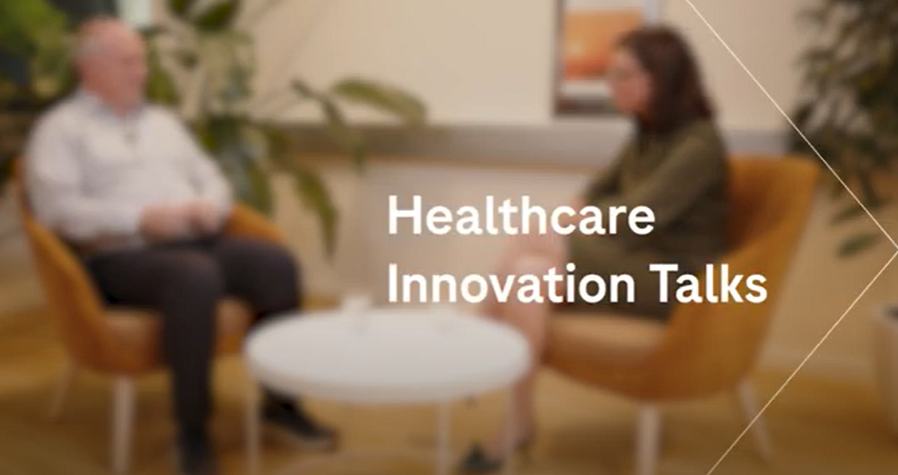}}
    % \dummyimage{3.2cm} % Uncomment for testing without actual image files
    \subcaption{Healthcare Innovation Talks}
  \end{subfigure}\hfill
  \begin{subfigure}[t]{0.32\textwidth}
    \centering
    \includegraphics[height=3.2cm, width=\linewidth]{{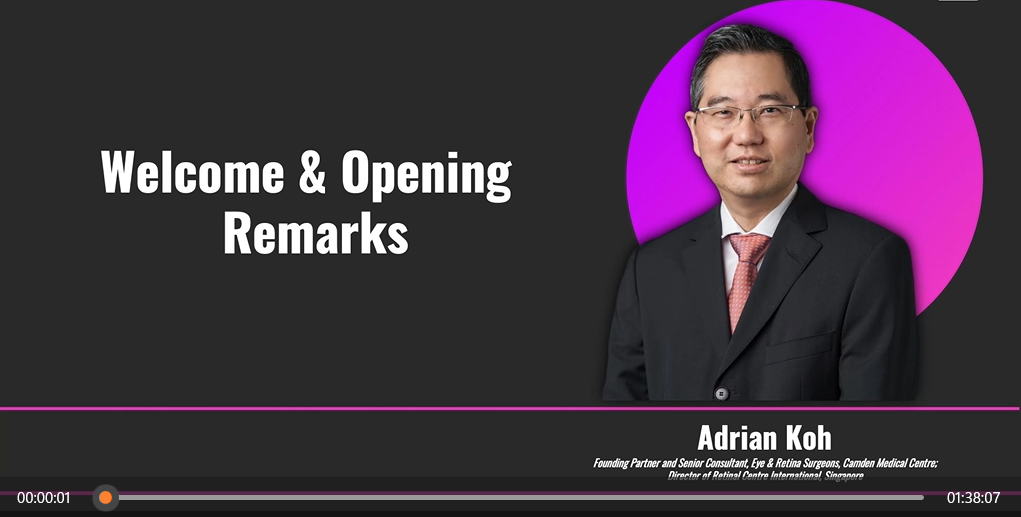}}
    % \dummyimage{3.2cm} % Uncomment for testing without actual image files
    \subcaption{Medical Conference Retro}
  \end{subfigure}\hfill
  \begin{subfigure}[t]{0.32\textwidth}
    \centering
    \includegraphics[height=3.2cm, width=\linewidth]{{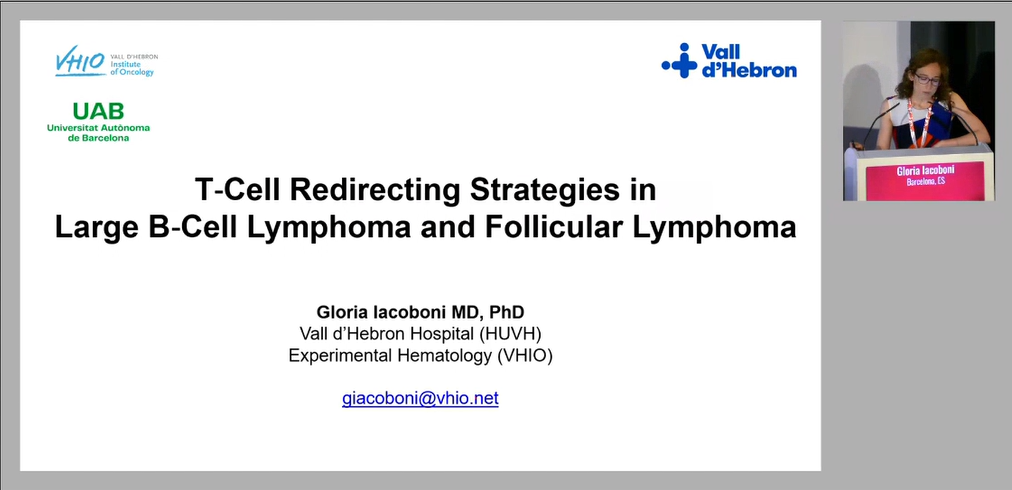}}
    % \dummyimage{3.2cm} % Uncomment for testing without actual image files
    \subcaption{Medical Forum}
  \end{subfigure}

  \vspace{6pt} % Provides a small vertical space between rows

  % ---------- Row 2 ----------
  \begin{subfigure}[t]{0.32\textwidth}
    \centering
    \includegraphics[height=3.2cm, width=\linewidth]{{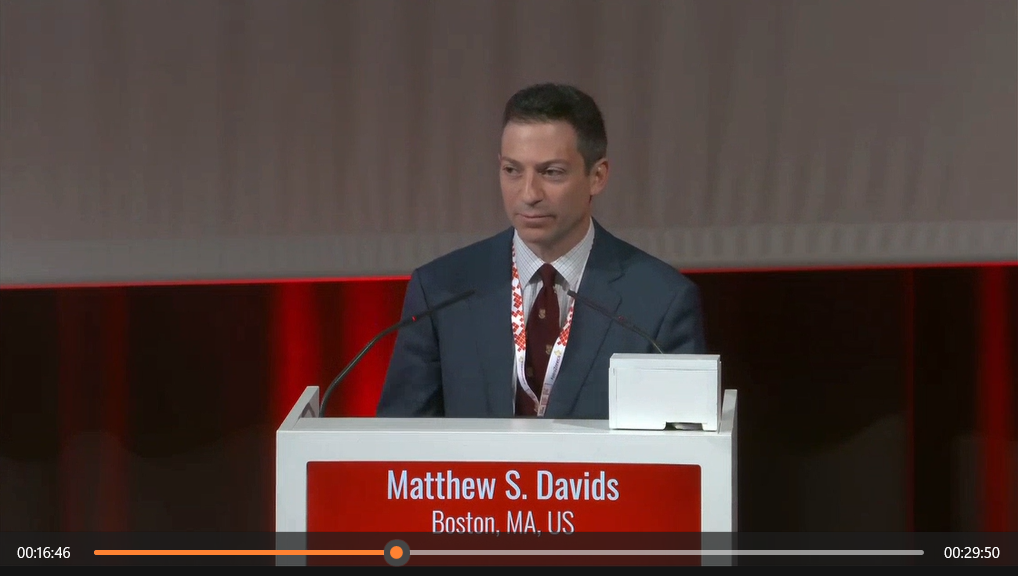}}
    % \dummyimage{3.2cm} % Uncomment for testing without actual image files
    \subcaption{Podium Speech}
  \end{subfigure}\hfill
  \begin{subfigure}[t]{0.32\textwidth}
    \centering
    \includegraphics[height=3.2cm, width=\linewidth]{{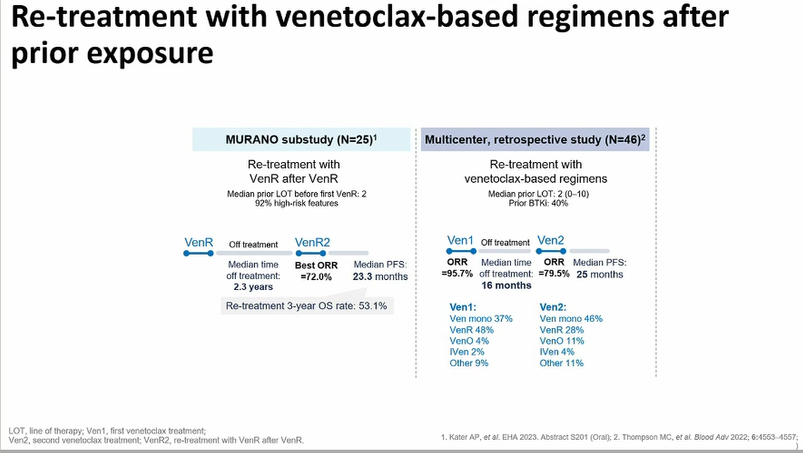}}
    % \dummyimage{3.2cm} % Uncomment for testing without actual image files
    \subcaption{Treatment Notes}
  \end{subfigure}\hfill
  \begin{subfigure}[t]{0.32\textwidth}
    \centering
    \includegraphics[height=3.2cm, width=\linewidth]{{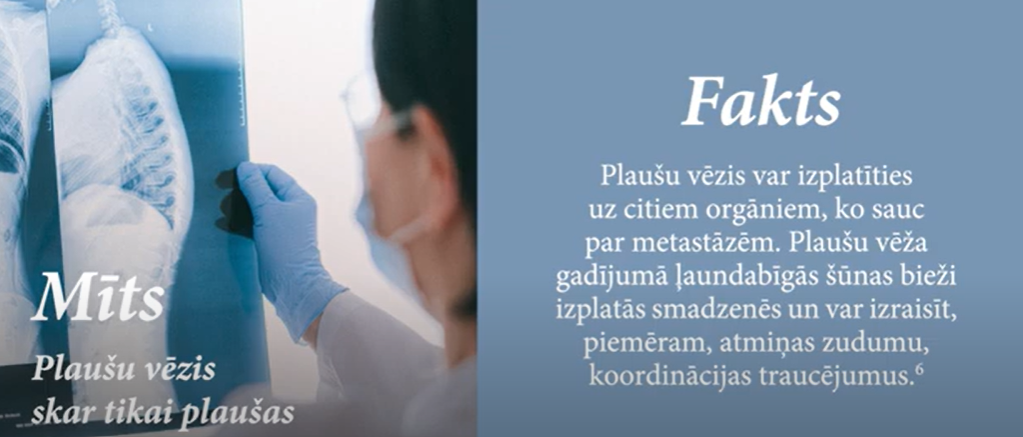}}
    % \dummyimage{3.2cm} % Uncomment for testing without actual image files
    \subcaption{Educational / Promotional Clip}
  \end{subfigure}

  \caption{\textbf{Sampled video inputs}: medical interview and lecture styles processed by our pipeline (talk-show interview, keynote remark, forum lecture, podium speech, treatment-slide session, and educational/promo clips).}
  \label{fig:expert-interviews-grid}
\end{figure*}

\begin{figure*}[htbp]
    \centering
    \begin{minipage}{\textwidth}
        \centering
        \includegraphics[width=1\linewidth]{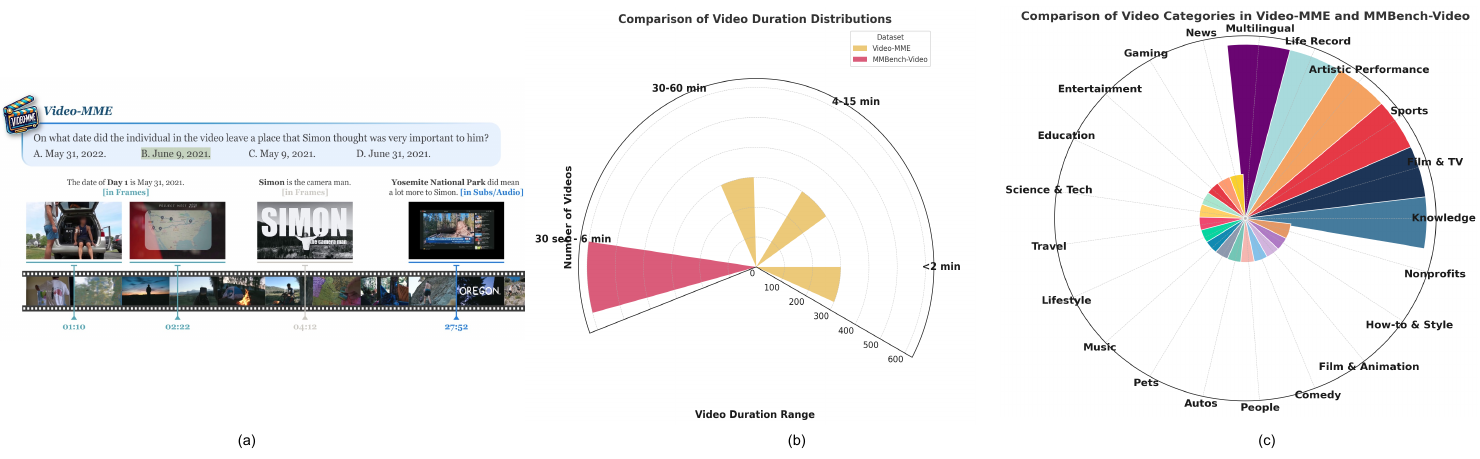}  
    \end{minipage}%
    \hfill
   
    \caption{\textbf{Comparison of the Video-MME \citep{fu2024video} and MMBench-Video datasets \citep{MMBench} in terms of video categories and duration distributions.} The Video-MME dataset consists of 900 videos spanning six primary visual domains with 30 subfields, categorized into 300 short-term (<2 min), 300 medium-term (4-15 min), and 300 long-term (30-60 min) videos.} 
    \label{fig: BenchmarkData}
\end{figure*}

\begin{figure*}[htbp]
  \centering
  % first row of three
  
    \centering
    \includegraphics[width=0.8\linewidth]{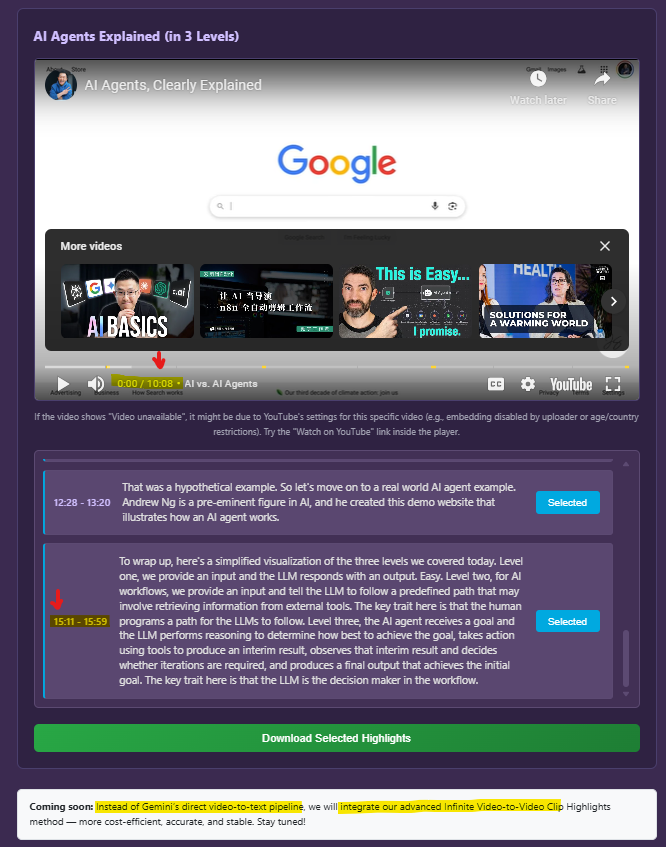}
       % optional
  \vspace{1ex} % small gap between rows
  \caption{\textbf{Gemini 2.5 's direct video-to-text feature still struggles with accurate timestamps } — for example, in a 10-minute video, it incorrectly selects a timestamp at the 15-minute mark.}
  \label{fig:failedcase}
\end{figure*}

In this section we provide the supplementary compiled together with the main paper includes:

Our Proprietary video dataset (<2 min, up to 3 h) distribution by scientific area, as shown in Table  \ref{tab:long-videos-areas}; The VideoMME Dataset, detailed in Table~\ref{fig: BenchmarkData}; Personalization and Key Segments Logic, along with sample outputs in Figure~\ref{fig:demo},~\ref {fig: thinkingchainoferrorcase} and Table ~\ref {tab: personalition}; Our video Clips landing scenarios on medical domain in Figure~\ref{fig:expert-interviews-grid}, Table~\ref {tab:phesgo_appendix}; Ablation Study on exciting video clip generation SaaS solutions, presented in Figure~\ref{fig:BenchmarkData 7}, and Tables~\ref{tab:bench-api}, ~\ref{tab:bench-keyframe} ~\ref{tab:bench-text}, ~\ref{tab:bench-git}; Gemini Video-to-Text Limitations, including issue analysis and failure case examples in Figure~\ref{fig:failedcase}, the cost breakdown in Table~ \ref{tab: cost}; Prompt Instruction ~\ref{tab:prompt_template}.

\begin{table}[ht]
\centering
\small
\caption{\textbf{Proprietary video dataset ($<$2\,min, up to 3\,h) distribution by scientific area}. Video includes clinical trials, interviews, medical lectures, promotional/non-promotional drug materials (tutorials, advertising), oral presentations, disease case studies, medical imaging/genomics demonstrations (CT scans, microscopy, DNA/sequencing), cancer morphology animations, and public health/disease education.}

\label{tab:long-videos-areas}
 \resizebox{0.50\textwidth}{!}{
\begin{tabular}{l r @{\hspace{1.5em}} l r}
\toprule
\multicolumn{2}{c}{\textbf{Scientific Area (A)}} & \multicolumn{2}{c}{\textbf{Scientific Area (B)}} \\
\cmidrule(lr){1-2}\cmidrule(lr){3-4}
\textbf{Area} & \textbf{Count} & \textbf{Area} & \textbf{Count} \\
\midrule
Oncology                & \textbf{5824} &  Movement Disorder      & 263           \\
Neuroscience           & \textbf{3306}   &  Nephrology             & 232  \\ 
Hematology             & \textbf{2369} & Infectious Disease     & 155 \\
Not Applicable         & 1556          & Inflammatory Disease   & 146  \\
Ophthalmology          & 1664          & Dermatology           & 30   \\
Respiratory Disease    & 312           & Cardiovascular         & 14 \\
Immunology             & 282           & Metabolism             & 6  \\

\bottomrule
\end{tabular}}
\end{table}

%%Qualitative Comparison of our Infinite Video-to-Infinite Video pipeline versus single-image-to-video baselines and commercial tool, shown in Figure~~\ref{fig:BenchmarkData 2}

          % space between figure and its caption

\begin{figure*}[ht]   % top of page to avoid large gaps
  \centering
  % ---- Row 1 ----
  \begin{subfigure}[t]{0.32\textwidth}
    \centering
    \includegraphics[width=\linewidth,height=48mm,keepaspectratio]{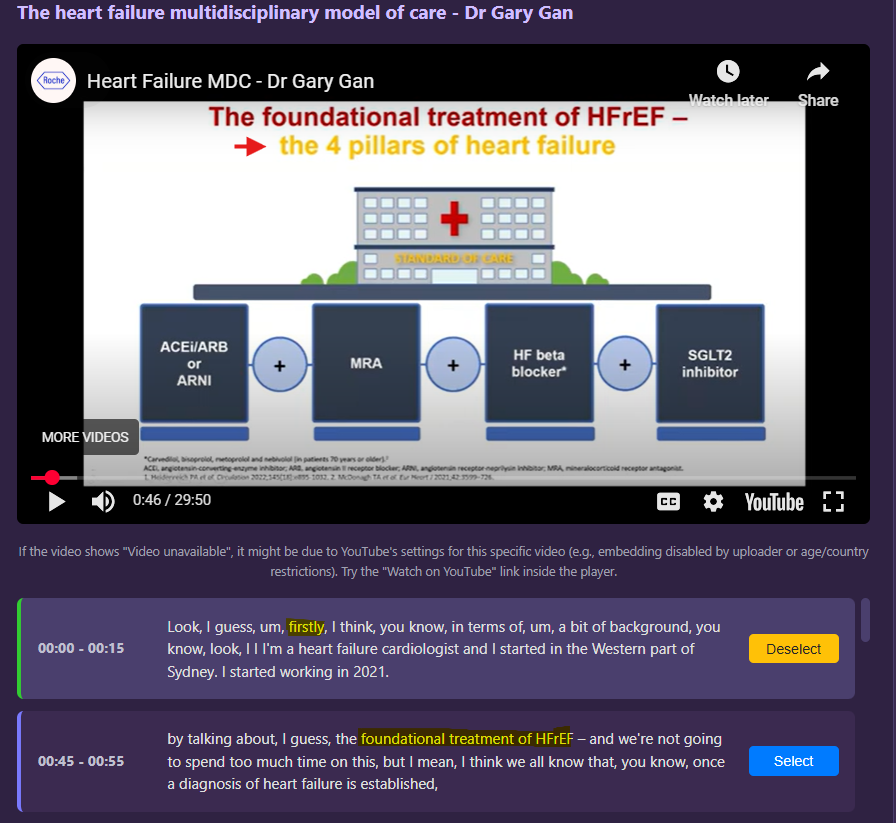}
    \caption{Agenda}
  \end{subfigure}\hfill
  \begin{subfigure}[t]{0.32\textwidth}
    \centering
    \includegraphics[width=\linewidth,height=48mm,keepaspectratio]{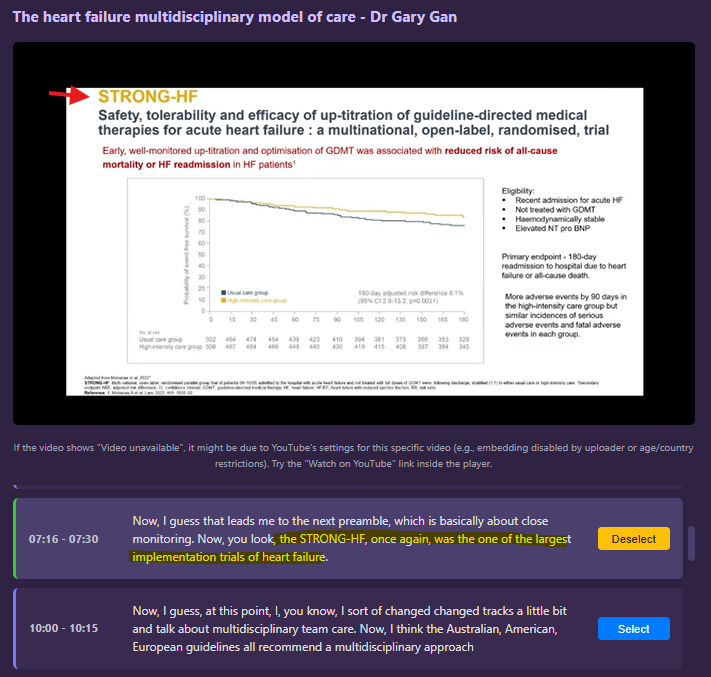}
    \caption{Keywords}
  \end{subfigure}\hfill
  \begin{subfigure}[t]{0.32\textwidth}
    \centering
    \includegraphics[width=\linewidth,height=48mm,keepaspectratio]{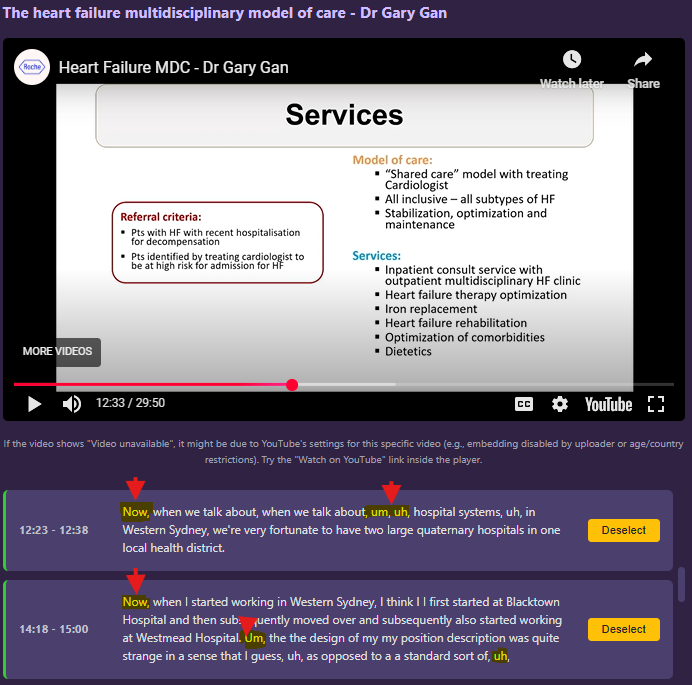}
    \caption{Speaker tone}
  \end{subfigure}

  \caption{\textbf{Impact of role definition and prompt-selection metrics.}  Agenda
and keyword alignment ensure coverage of key sections; speaker tone, transitions, and fade-in/out reduce visual–voice mismatch and jump cuts; full-length coverage prevents unfinished and fragmented clips.}
  \label{fig:personalization}
\end{figure*}

\begin{table*}[t]
\centering
\small
\begin{tabular}{p{0.22\linewidth} p{0.7\linewidth}}
\toprule
\rowcolor{rocheblue!20}\textbf{Landing scenarios} & \textbf{Patients/Doctors Interviews Listening} \\
\midrule
\textbf{Consolidate summary across interviews}\textit{
- Q1: Consolidate summaries for each interview quickly and cost-effectively via our pipeline?} 
& 
\textbf{Dr. T:} Presents a study showing low rates of hypersensitivity with Phesgo, supporting its home administration. She outlines the benefits of home administration for patient quality of life and identifies suitable patient populations and implementation challenges. \newline
\textbf{Dr. C:} Provides instructions on preparing and administering Phesgo as a subcutaneous injection, highlighting steps like gathering supplies, cleaning, drawing the correct dose, and patient monitoring. \newline
\textbf{Dr. V:} Discusses the development of subcutaneous drugs like Herceptin and Phesgo, emphasizing improved patient experience by moving administration out of hospitals to more convenient locations. He also covers dosing, safety, and training. \newline
\textbf{Dr. L:} Discusses the AFFINITY trial, which demonstrated an overall survival benefit for adding pertuzumab to trastuzumab and chemotherapy in HER2+ early breast cancer patients with higher risk factors, particularly those with node involvement.. \\
\midrule
\textbf{Track who said what at when}\textit{ -
Q2: Would you prefer to use Phesgo in the neoadjuvant setting and in which patient population, or in adjuvant Residual Disease?} 
&
\textbf{Dr. T:} mentions its use in the adjuvant setting after completion of chemotherapy for a year of therapy \textcolor{rocheblue}{(segments 40-43, approx. 3:12.12-3:28.12)}. She also describes its use after pre-operative (neoadjuvant) therapy if a patient achieves a PCR and needs to complete a year of HER2-targeted therapy \textcolor{rocheblue}{(segments 48-49, approx. 3:38.64-3:48.52)}. She states, 'In the preoperative setting, we're usually giving chemo with PheSGO so that wouldn't make sense [for home administration as they are coming in for chemo]' \textcolor{rocheblue}{(segments 45-47, approx. 3:31.64-3:38.64)} \newline
\textbf{Dr. C:} mentions his unit has been using 'FESCO' (Phesgo) 'for the last four years in the metastatic neoadjuvant and adjuvant setting' \textcolor{rocheblue}{(segment 2, 0:20.52-0:25.76)}. This confirms its use in these settings but his interview focuses on the practical administration rather than clinical decision-making or preference between neoadjuvant and adjuvant residual disease, or specific patient populations beyond general HER2-positive eligibility. \newline
\textbf{Dr. V:} discusses his center's involvement with 'FESGO' (Phesgo) starting with the Federica trial and rapid transition to its use once commercially available \textcolor{rocheblue}{(segments 5-8, approx. 0:41.58-1:06.76)}. He mentions giving Phesgo with chemotherapy \textcolor{rocheblue}{(segment 53, approx. 5:43.44-5:51.02: 'because they're getting chemotherapy at the same time patients are usually in the hospital when they get the first doses of subcut treatment')}, which is common in the neoadjuvant setting and potentially for concurrent adjuvant chemotherapy. He also discusses patients continuing treatment after chemotherapy, fitting the adjuvant setting. He states, 'most people having chemotherapy sitting in a chair' \textcolor{rocheblue}{(segment 44, approx. 4:57.9-5:05.18) in the context of identifying patients for Phesgo}. \newline
\textbf{Dr. L:} focuses entirely on the AFFINITY trial results, which investigated the addition of (IV) pertuzumab to trastuzumab and chemotherapy. \textcolor{rocheblue}{Phesgo is not mentioned in her transcript}. \\
\midrule
\textbf{Key theme per molecules, trial/drugs}\textit{-
Q3: In light of upcoming Destiny Breast-09 readout, do you believe that all patients would need Phesgo till progression or would you consider induction Phesgo followed by Perjeta - Herceptin maintenance strategy, for which patients and what data you would need to see to implement that} 
& 
\textbf{Dr. T:} \textcolor{rocheblue}{does not} explicitly state a *preference* for neoadjuvant vs. adjuvant residual disease, but describes its current use in both contexts where appropriate (adjuvant after chemo, or continuing HER2 therapy post-neoadjuvant chemo/surgery if PCR achieved). \newline
\textbf{Dr. C:} mentions his unit has been using 'FESCO' (Phesgo) 'for the last four years in the metastatic neoadjuvant and adjuvant setting' (segment 2, 0:20.52-0:25.76). \textcolor{rocheblue}{This confirms its use in these settings but his interview focuses on the practical administration rather than clinical decision-making or preference between neoadjuvant and adjuvant residual disease, or specific patient populations beyond general HER2-positive eligibility.} \newline
\textbf{Dr. V:} \textcolor{rocheblue}{doesn't express a specific preference }for neoadjuvant versus adjuvant residual disease for Phesgo itself, nor does he detail specific patient population criteria beyond HER2-positivity. His focus is on decentralization and patient experience. \newline
\textbf{Dr. L:}  interview \textcolor{rocheblue}{focuses entirely on the AFFINITY trial results}, which investigated the addition of (IV) pertuzumab to trastuzumab and chemotherapy. \textcolor{rocheblue}{Phesgo is not mentioned} in her transcript. \\
\bottomrule
\end{tabular}
\caption{\textbf{Landing scenario: Structured medical interview Q\&A output from our pipeline.} 
Given multiple expert interviews, the pipeline extracts, identifies, and consolidates opinions around specific clinical queries (e.g., Phesgo usage). 
As queries become more domain-specific and knowledge-intensive, our system effectively handles such cases with precise localization of quoted statements. 
This demonstrates how structured summarization can support rapid evidence synthesis from long-form medical interviews.}

\label{tab:phesgo_appendix}
\end{table*}

% ==============================================================================
% TABLE 1: AI Video Generators from Video API (Commercial)
% ==============================================================================
\begin{table*}[h]
  \centering
  \small
  \renewcommand{\arraystretch}{1.2}
  \setlength{\tabcolsep}{8pt}
  \caption{AI Video Generators from Video API (Commercial)}
  \label{tab:bench-api}
  \begin{tabularx}{\textwidth}{@{} l|X @{}}
    \toprule
    \rowcolor{rocheblue!20}
    \textbf{Company / Startup} & \textbf{Use case \& Key features} \\
    \midrule

    \href{https://www.aistudios.com/}{DeepBrain API}
      & Talking avatars; stylized visuals for short reels. \\

    \addlinespace
    \href{https://runwayml.com}{Runway API}
      & Raw video generation from text prompts. \\

    \addlinespace
    \href{https://elevenlabs.io/}{ElevenLabs + D-ID}
      & Generate audio and sync to an avatar face; entertainment-industry grade. \\

    \addlinespace
    \href{https://stability.ai/stable-video}{Stability AI Sora}
      & Image-to-video with full-scene 3D generation. \\

    \bottomrule
  \end{tabularx}
\end{table*}

% ==============================================================================
% TABLE 2: AI Video Generators from Keyframe Images (Commercial)
% ==============================================================================
\begin{table*}[h]
  \centering
  \small
  \renewcommand{\arraystretch}{1.2}
  \setlength{\tabcolsep}{8pt}
  \caption{AI Video Generators from Keyframe Images (Commercial)}
  \label{tab:bench-keyframe}
  \begin{tabularx}{\textwidth}{@{} l|X @{}}
    \toprule
    \rowcolor{rocheblue!20}
    \textbf{Company / Startup} & \textbf{Use case \& Key features} \\
    \midrule

    \href{https://www.kaiber.ai}{Kaiber}
      & Turn images or music into animated videos; stylized visuals for music videos and short reels. \\

    \addlinespace
    \href{https://animoto.com}{Animoto}
      & Slideshow-style video maker with text overlays, captions, and voice-over to produce professional clips. \\

    \bottomrule
  \end{tabularx}
\end{table*}

% ==============================================================================
% TABLE 3: AI Video Generators from Text / Transcript (Commercial)
% ==============================================================================
\begin{table*}[h]
  \centering
  \small
  \renewcommand{\arraystretch}{1.2}
  \setlength{\tabcolsep}{8pt}
  \caption{AI Video Generators from Text / Transcript (Commercial)}
  \label{tab:bench-text}
  \begin{tabularx}{\textwidth}{@{} l|X @{}}
    \toprule
    \rowcolor{rocheblue!20}
    \textbf{Company / Startup} & \textbf{Use case \& Key features} \\
    \midrule

    \href{https://runwayml.com}{Runway Gen-2}
      & Text-to-video or transform still images into 5–10 s motion clips; very strong generative capability (e.g.\ “flying car from a static image”). \\

    \addlinespace
    \href{https://www.pika.art}{Pika Labs}
      & Generate short cinematic clips from prompts or stills; excellent for storytelling and scene/character animation. \\

    \addlinespace
    \href{https://www.synthesia.io}{Synthesia}
      & Create talking-head videos from scripts, with avatars speaking multiple languages—ideal for explainers and tutorials. \\

    \addlinespace
    \href{https://www.heygen.com}{HeyGen}
      & Avatar-based video generation from transcripts; high-quality avatars with realistic lip sync. \\

    \addlinespace
    \href{https://www.veed.io}{Veed.io}
      & Video editing plus AI generation and stock templates; combines text, images, voice over, and stock clips. \\

    \bottomrule
  \end{tabularx}
\end{table*}

% ==============================================================================
% TABLE 4: AI Video Generators from Open Source Git (Open-source)
% ==============================================================================
\begin{table*}[h]
  \centering
  \small
  \renewcommand{\arraystretch}{1.2}
  \setlength{\tabcolsep}{8pt}
  \caption{AI Video Generators from Open Source Git (Open-source)}
  \label{tab:bench-git}
  \begin{tabularx}{\textwidth}{@{} l|X @{}}
    \toprule
    \rowcolor{rocheblue!20}
    \textbf{Git tool} & \textbf{Use case \& Key features} \\
    \midrule

    \href{https://github.com/AILab-CVC/StoryDiffusion}{Story Diffusion}
      & Generate a short story video from a script: Script → Scenes → Image generation → Video stitching. \\

    \addlinespace
    \href{https://github.com/Rudrabha/Wav2Lip}{OpenVPI / Wav2Lip}
      & Sync lip movements to speech by taking a face image + audio and producing a synced talking face. \\

    \addlinespace
    \href{https://genmo.ai}{Genmo AI (free tier + API)}
      & Generate stylized videos from text or stills; partially open/research-based access. \\

    \addlinespace
    \href{https://github.com/deforum-art/deforum-stable-diffusion}{Deforum}
      & Create motion videos from prompts or image sequences, built on Stable Diffusion with animation configs. \\

    \addlinespace
    \href{https://github.com/OpenTalk-AI/OpenTalk}{OpenTalk / SadTalker}
      & Educational or narrator-style talking-face generation from a still image + audio. \\

    \bottomrule
  \end{tabularx}
\end{table*}

\begin{table*}[htbp]
  \centering
  \small
  \renewcommand{\arraystretch}{1.2} % slightly taller rows
  \setlength{\tabcolsep}{8pt}       % horizontal padding

  \caption{\textbf{Gemini 2.5 Pro Multimodal Reasoning Chain }for \textcolor{rocheblue}{Fail-Case} Video ID: vDzNVHXc66U from Video-MME. It demonstrates the logic behind Gemini 2.5 Pro's processing steps after a video upload. Notably, extensive metadata retrieval via the Google Search API or related services occurs even before Gemini really "watching" the video content, as highlighted in blue.}
  \label{tab:gemini-chain}

  \begin{tabularx}{\textwidth}{@{} l|X @{}}
    \toprule
    \rowcolor{rocheblue!20}
    \textbf{Step} & \textbf{Description} \\
    \midrule

    Reframing the Approach
      & I'm now shifting my focus to 
        \textcolor{rocheblue}{analyzing the video itself}, 
        as the transcript proved unhelpful. The user's prompt suggests a 
        direct video-analysis approach is required to create an effective 
        highlight reel. \\

    \addlinespace
    Adjusting the Strategy
      & I've realized that my previous approach, despite using 
        \textcolor{rocheblue}{\texttt{use\_audio\_video\_tokens=True}}, 
        didn't adequately focus on the visual content. My initial prompt was 
        still too centered on spoken words. \\

    \addlinespace
    Implementing Visual Analysis
      & The previous try was ineffective due to the useless transcript. Now, 
        I'm concentrating on the video's visual elements: requesting 
        start/end times and scene descriptions, then formatting the output as 
        JSON with visual cues in place of spoken text. \\

    \addlinespace
    Exploring Visual Content Extraction
      & I'm currently trying to extract key segments from the video based on 
        visual content. My 
        \textcolor{rocheblue}{initial strategy with YouTube failed}, 
        so I'm exploring object-detection and scene-recognition models to 
        pinpoint visually important elements. \\

    \addlinespace
    Assessing Metadata Retrieval
      & I've hit a 
        \textcolor{rocheblue}{roadblock with direct visual analysis due to a 
        TypeError when using the YouTube tool}. I'm pivoting to 
        \textcolor{rocheblue}{retrieving video metadata via YouTube}. \\

    \addlinespace
    Refocusing on Metadata
      & After the TypeError, I re-examined the problem and refocused on 
        \textcolor{rocheblue}{retrieving video metadata via YouTube}. 
        While \textcolor{rocheblue}{not ideal}, it’s the only remaining path 
        given the current tool limitations. \\

    \addlinespace
    Investigating Video Content
      & I'm still trying to extract key visual segments. Both transcript and 
        metadata strategies have failed, so I’m 
        \textcolor{rocheblue}{exploring alternative APIs and CV toolkits} 
        to process the raw frames directly. \\

    \addlinespace
    Reframing Approach to the Video
      & I re-evaluated the YouTube tool’s requirements: it needs a valid 
        \textcolor{rocheblue}{\texttt{video\_id}}. The provided URL https://www.youtube.com/watch?v=vDzNVHXc66U may be a 
        placeholder, so I’m testing with a dummy ID (‘0’) to probe the tool’s 
        behavior. \\

    \addlinespace
    Pivoting Strategy for Video ID
      & Recognizing the unusual link, I’m now assuming ‘0’ is the ID and calling 
        YouTube again to retrieve title/description and verify whether any 
        metadata is accessible. \\

    \addlinespace
    Facing Technical Roadblocks
      & Critical tool failures persist, halting any attempt to identify visual 
        segments. I’m investigating workarounds but the issues remain unresolved. \\

    \addlinespace
    Pinpointing Analysis Limitations
      & I’ve 
        \textcolor{rocheblue}{hit a wall: the audio transcript is useless gibberish, visual-analysis calls error out, and metadata retrieval fails}. \\

    \addlinespace
    Identifying Processing Barriers
      & I’m 
        \textcolor{rocheblue}{unable to process the video or extract information}. 
        Neither transcript, CV pipelines, nor metadata calls succeed. I cannot 
        fulfill the request at this time. \\

    \bottomrule
  \end{tabularx}
  \label{fig: thinkingchainoferrorcase}
\end{table*}

\begin{table*}[htbp]
  \centering
  \small
  \renewcommand{\arraystretch}{1.2}
  \setlength{\tabcolsep}{8pt}
  \caption{\textbf{Personalization via Gemini 2.5 Pro Multimodal Reasoning}:
Side-by-side showing Gemini “thinking chains” for generating marketing-promo vs. training videos clips, showing how transcript segments are selected and formatted into JSON for clip generation.}
  \label{tab:gemini-personalization}
  \begin{tabularx}{\textwidth}{@{} l|X|X @{}}
    \toprule
    \rowcolor{rocheblue!20}
    \textbf{Step} 
      & \textbf{Marketing/Promo Pipeline} 
      & \textbf{Training/Instruction Pipeline} \\
    \midrule

    Reframing the Approach
      & \textcolor{rocheblue}{Focus on brand hooks and emotional leads}:  
        “Which moments sell the product benefit?”  
      & \textcolor{rocheblue}{Focus on concept clarity and tool intros}:  
        “Which segments clearly explain ‘what and why’?” \\

    \addlinespace
    Adjusting the Strategy
      & Emphasize \textcolor{rocheblue}{calls-to-action, upbeat music cues, logo reveals}.  
      & Emphasize \textcolor{rocheblue}{step-by-step demos, key terminology definitions, “do’s don’ts.” }\\

    \addlinespace
    Segment Selection
      & Pick \textcolor{rocheblue}{high-impact visuals: product shots, testimonials, USPs}.  
      & Pick \textcolor{rocheblue}{explanatory visuals: UI walkthroughs, process flows, compliance notes.} \\

    \addlinespace
    JSON Formatting
      & Output as promo video segments with  
        start/end/text JSON for ad-style snippets.  
      & Output as training video segments with  
        start/end/text JSON for tutorial modules. \\

    \addlinespace
    Immersive Update
      & Update the existing immersive (`id="promo video segments"`)  
        to \textcolor{rocheblue}{focus on marketing highlights}.  
      & If persona changes, replace that immersive with  
        `id="training video segments"` \textcolor{rocheblue}{containing tutorial clips}. \\

    \bottomrule
  \end{tabularx}
  \label{tab: thinkingchain}
\end{table*}

\begin{table*}[htbp]
  \centering
  \small
  \renewcommand{\arraystretch}{1.2}
  \setlength{\tabcolsep}{8pt}
  \caption{Comparison of Paid Tier Pricing for Gemini 2.5 Models (per 1M tokens in USD)}
  \label{tab:gemini_pricing}
  \begin{tabularx}{\textwidth}{@{} l|X|X|X @{} }
    \toprule
    \rowcolor{rocheblue!20}
    \textbf{Pricing (USD per 1M tokens)} & \textbf{Gemini 2.5 Pro} & \textbf{Gemini 2.5 Flash} & \textbf{Gemini 2.5 Flash-Lite Preview} \\
    \midrule

    Input Price & \textcolor{rocheblue}{$1.25$ (< 200k tokens), $2.50$ (> 200k tokens)} & \textcolor{rocheblue}{$0.30$ (text/image/video), $1.00$ (audio)} & \textcolor{rocheblue}{$0.10$ (text/image/video), $0.50$ (audio)} \\

    \addlinespace
    Output Price & \textcolor{rocheblue}{$10.00$ (< 200k tokens), $15.00$ (> 200k tokens)} & \textcolor{rocheblue}{$2.50$} & \textcolor{rocheblue}{$0.40$} \\

    \addlinespace
    Context Caching & \textcolor{rocheblue}{$0.31$ (< 200k tokens), $0.625$ (> 200k tokens), $4.50$ / 1M tokens} & \textcolor{rocheblue}{$0.075$ (text/image/video), $0.25$ (audio), $1.00$ / 1M tokens} & \textcolor{rocheblue}{$0.025$ (text/image/video), $0.125$ (audio), $1.00$ / 1M tokens} \\

    \addlinespace
    Grounding with Google Search & 1,500 RPD (free), then \$35 / 1,000 requests & 1,500 RPD (free), then \$35 / 1,000 requests \& 1,500 RPD (free), then \$35 / 1,000 requests \\

    \bottomrule
  \end{tabularx}
  \label{tab: cost}
\end{table*}

\begin{table*}[htbp]
  \centering
  \small
  \renewcommand{\arraystretch}{1}
  \setlength{\tabcolsep}{2pt}
  \caption{\textbf{Personalized Video Clips Comparison of Selected Segments for Marketing/Promo vs. Training/Educational Clips } based on same input \textcolor{red}{10mins} Youtube video: \href{https://www.youtube.com/watch?v=FwOTs4UxQS4/}{FwOTs4UxQS4} using \textcolor{red}{Gemini 2.5 Flash ( Direct Video to Text)}. Different goals call for different segment selections: Marketing clips emphasize high-impact openings, motivational statements, famous expert mentions, pro tips, and distinctive traits of AI agents—highlighted in \textcolor{rocheblue}{blue}. In contrast, training clips prioritize a structured explanation of concepts, such as the three-level AI framework, detailed characteristics of LLMs and workflows, hypothetical examples. Here, Gemini 2.5 still struggles with generating correct timestamps, as highlighted in \textcolor{red}{red}.}
  \label{tab:segment_comparison}
  \resizebox{0.8\textwidth}{!}{
  %\begin{adjustbox}{max height=0.5\textheight, center}
  \begin{tabularx}{\textwidth}{@{} l|X|X @{} }
    \toprule
    \rowcolor{rocheblue!20}
     \textbf{Timestamp (s)} & \textbf{Marketing/Promotion Clips} & \textbf{Training/Educational Clips} \\
    \midrule

    0.2--17.1 & ... \textcolor{rocheblue}{the most important sentence in this entire video}, the one massive change that has to happen in order .....the human decision maker, to be replaced by an LLM. & -- \\

    \addlinespace
    19.1--23.9 & ... most explanations of AI agents is either too technical or too basic. & ... most explanations of AI agents is either too technical or too basic. \textcolor{rocheblue}{This video is meant for people like myself.} \\

    \addlinespace
    34.4--48.4 & ...No matter how many steps we add, this is \textcolor{rocheblue}{still just an AI workflow...} & -- \\

    \addlinespace
    38.4--55.2 & ...follow a \textcolor{rocheblue}{ simple 1-2-3 learning path by building on concepts you already understand}, like ChatGPT, and then moving on to AI workflows, and then finally AI agents. \textcolor{rocheblue}{All the while using examples you'll actually encounter in real life.} & -- \\

    \addlinespace
    49.3--52.0 & \textcolor{rocheblue}{Pro tip}: Because of this, the most common configuration for AI agents is the ReAct framework... & -- \\

    \addlinespace
    3.2--22.3 & ... \textcolor{rocheblue}{key trait} of AI agents is their ability to iterate.... rewrite the prompt to make the LinkedIn post funnier? ...the human,..repeat this iterative process a few times to get something & -- \\

    \addlinespace
    48.2--63.4 & ... real world AI agent example. \textcolor{rocheblue}{Andrew Ng} is a pre-eminent figure in AI, ... created this demo website that illustrates how an AI agent works... And then it's acting by looking at clips in video footage, ...indexing that clip, and then returning that clip to us. & -- \\

    \addlinespace
    36.3--59.5 & Level three: AI agents. The AI agent receives a goal and the LLM performs reasoning to determine how best to achieve the goal,... \textcolor{rocheblue}{The key trait here is that the LLM is the decision maker in the workflow.} & -- \\

    \addlinespace
    107.5--119.0 & -- & \textcolor{rocheblue}{Kicking things off at level one}, large language models. Popular AI chatbots like ChatGPT,...are applications built on top of large language models, LLMs, ...fantastic at generating and editing text. \\

    \addlinespace
    157.0--214.0 & -- & ... \textcolor{rocheblue}{two key traits of large language models}. First, despite being trained on vast amounts of data, they have limited knowledge of proprietary information: like our personal information or internal company data. Second, LLMs are passive - they wait for our prompt and then respond,.. \\

    \addlinespace
    218.0--222.0 & -- & \textcolor{rocheblue}{Moving to level two}, AI workflows. Let's build on our example... \\

    \addlinespace
    304.0--316.0 & -- & This is a fundamental trait of AI workflows. They can only follow predefined paths set by humans... this path is also called the control logic. \\

    \addlinespace
    527.0--534.0 & -- & \textcolor{rocheblue}{All right, level three}, AI agents. Continuing the make.com example... \\

    \addlinespace
    600.0--617.0 & ... this is the most important sentence in this entire video, the one massive change that has to happen in order for this AI workflow to become an AI agent is for me, the human decision maker, to be replaced by an LLM. &...and this is the most important sentence in this entire video, ...the human decision maker, to be replaced by an LLM. \\

    \addlinespace
     \textcolor{red}{649.0--702.0} & Pro tip: ...the most common configuration for AI agents is the ReAct framework. All AI agents must reason and act, so ReAct. & Pro tip: Because of this, the most common configuration for AI agents is the ReAct framework. ..., so ReAct. \textcolor{rocheblue}{Sound simple once we break it down, right?} \\

    \addlinespace
     \textcolor{red}{703.0--722.0} & -- & A third key trait of AI agents is their ability to iterate. Remember when I had to manually rewrite the prompt to make the LinkedIn post funnier? I, the human, probably need to repeat this iterative process a few times to get something I'm happy with... \\

    \addlinespace
     \textcolor{red}{748.0--800.0} & -- & \textcolor{rocheblue}{That was a hypothetical example}... move on to a real world AI agent example. Andrew Ng is a pre-eminent figure in AI, and he created this demo website that illustrates how an AI agent works. \\

    \addlinespace
     \textcolor{red}{911.0--959.0} & -- & \textcolor{rocheblue}{To wrap up}, ...the three levels ... Level one, ... input and the LLM responds with an output...Level two, .. input and tell the LLM to follow a predefined path ... The key trait ... the human programs a path for the LLMs to follow. Level three, the AI agent receives a goal ... LLM is the decision maker in the workflow. \\

    \bottomrule
  \end{tabularx}}
  \label{tab: personalition}
\end{table*}

\begin{table*}[ht]
\centering
\caption{structured Role-Based Prompt for Extractive Video Highlight Selection}
\begin{tabular}{p{0.20\linewidth} p{0.8\linewidth}}
\toprule
\rowcolor{rocheblue!20}\textbf{System Prompt} & \textbf{Content} \\
\midrule

\textbf{System Role} &
You are an expert assistant for selecting the most meaningful content from a video.
Your task is to identify and extract important segments that together form a highlight of up to 3 minutes.
Use the original spoken text exactly as-is. Do not paraphrase. \\

\midrule
\textbf{Task Overview} &
When a YouTube video URL is provided for direct video analysis, segment selection must be derived from the \textbf{actual visual and audio content of the video}.
For other URLs or general topics, analysis should be based on understanding the provided material.
The final segments must correspond to content that could be directly extracted from \textbf{the original video}. \\

\midrule
\textbf{Segment Selection Criteria} &
\begin{itemize}\itemsep0pt
\item Reflect the most important ideas, agenda points, or transitions.
\item Ensure coverage across the full video duration, including beginning and end sections when relevant.
\item Consider speaker tone, pauses, and natural breaks to ensure smooth clip transitions.
\end{itemize} \\

\midrule
\textbf{Critical Instruction} &
\textbf{You MUST preserve the exact wording, phrasing, and sentences from the original video. 
Do not rephrase, summarize, or generate new text. 
All extracted text must be copied verbatim from the source.} \\

\midrule
\textbf{Output Format Requirements} &
Respond with a valid JSON object. The entire response must be correctly formatted and parsable. \\

\midrule
\textbf{Timestamp Rules} &
\begin{itemize}\itemsep0pt
\item Do not use timestamps in \texttt{HH:MM:SS} format.
\item Convert all time references into seconds only.
\item Use at most two decimal places for timestamps.
\item If timestamps are not applicable, use \texttt{"N/A"} or omit the \texttt{start} and \texttt{end} fields.
\end{itemize} \\

\midrule
\textbf{YouTube-Specific Constraint} &
For YouTube videos, when a URL is provided for direct analysis, the \texttt{title} field in the JSON output MUST be the original, exact title of the video.
Do not generate, summarize, or rephrase the title. \\

\midrule
\textbf{Required Output JSON Structure} &
\begin{verbatim}
  "select_segments": [
    {
      "start": 12.5,
      "end": 25.3,
      "text": "We begin ..."
    }
  ]
\end{verbatim}
\\

\midrule
\textbf{User Customization} & The following fields are injected dynamically at runtime:
\begin{itemize}\itemsep0pt
\item \textbf{User-Provided Role / Style}, e.g.,
\textit{``technical trainer for a marketing campaign''}.
\item \textbf{User-Provided Additional Requirements}, e.g.,
\textit{``focus on business benefits''}
\end{itemize} \\

\bottomrule
\end{tabular}
\label{tab:prompt_template}
\end{table*}

\end{document}